\RequirePackage{fix-cm}
\documentclass[twocolumn]{svjour3}          
\smartqed  
\usepackage{graphicx}
\usepackage[svgnames]{xcolor}
\usepackage{color, colortbl}
\usepackage{caption}
\usepackage{microtype}
\usepackage{times}
\usepackage{epsfig}
\usepackage{graphicx}
\usepackage{amsmath}
\usepackage{amssymb}
\usepackage{tikz}
\usetikzlibrary{shapes,arrows}
\usepackage{changepage}
\usepackage{algorithm}
\usepackage{algpseudocode}
\usepackage{multirow}
\usepackage{subcaption} 
\usepackage[breaklinks=true,draft]{hyperref}
\usepackage{breakcites}
\usepackage{natbib}

\newcommand\ie{\emph{i.e.}}
\newcommand\eg{\emph{e.g.}}

\newcommand\bb[1]{\textbf{#1}}

\definecolor{Gray}{gray}{0.9}

\pgfdeclarelayer{edgelayer}
\pgfdeclarelayer{nodelayer}
\pgfsetlayers{edgelayer,nodelayer,main}

\tikzstyle{none}=[inner sep=2pt]

\tikzstyle{vertex_default}=[circle,fill=White,draw=Black,text width=0.3cm]
\tikzstyle{vertex_hyper}=[ellipse,minimum width=50,minimum height=21,ultra thick,fill=White,draw=Green]
\tikzstyle{edge_forward}=[->,ultra thick]
\tikzstyle{edge_forward_selected}=[->,ultra thick,draw=Red]
\tikzstyle{edge_backward}=[<-,ultra thick,draw=Green]
\tikzstyle{edge_backward_selected}=[<-,ultra thick,draw=Blue]
\def\makeheadbox{\relax}

\begin{document}

\title{Learning Optimal Parameters for Multi-target Tracking with Contextual Interactions
\thanks{This work was supported by the US National Science Foundation through
awards IIS-1253538 and DBI-1053036}}



\author{Shaofei Wang         \and
        Charless C. Fowlkes 
}


\institute{S. Wang \at
              \email{sfwang0928@gmail.com}           
           \and
           C. Fowlkes \at
              Dept. of Computer Science\\
              University of California, Irvine\\
              \email{fowlkes@ics.uci.edu}           
}

\date{}

\maketitle

\begin{abstract}
We describe an end-to-end framework for learning parameters of min-cost flow
multi-target tracking problem with quadratic trajectory interactions including
suppression of overlapping tracks and contextual cues about co-occurrence of
different objects. Our approach utilizes structured prediction with a
tracking-specific loss function to learn the complete set of model parameters.
In this learning framework, we evaluate two different approaches to finding an
optimal set of tracks under a quadratic model objective, one based on an LP
relaxation and the other based on novel greedy variants of dynamic programming
that handle pairwise interactions. We find the greedy algorithms achieve almost
equivalent accuracy to the LP relaxation while being up to 10x faster than a
commercial LP solver.  We evaluate trained models on three challenging
benchmarks.  Surprisingly, we find that with proper parameter learning, our
simple data association model without explicit appearance/motion reasoning is
able to achieve comparable or better accuracy than many state-of-the-art methods
that use far more complex motion features or appearance affinity metric
learning.
\end{abstract}

\section{Introduction}
\label{sec:intro}
\begin{figure*}[t]
\begin{center}
\begin{tabular}{cc}
\includegraphics[clip,trim=0cm 0cm 0cm 0cm,width=0.47\textwidth,height=0.27\textwidth]{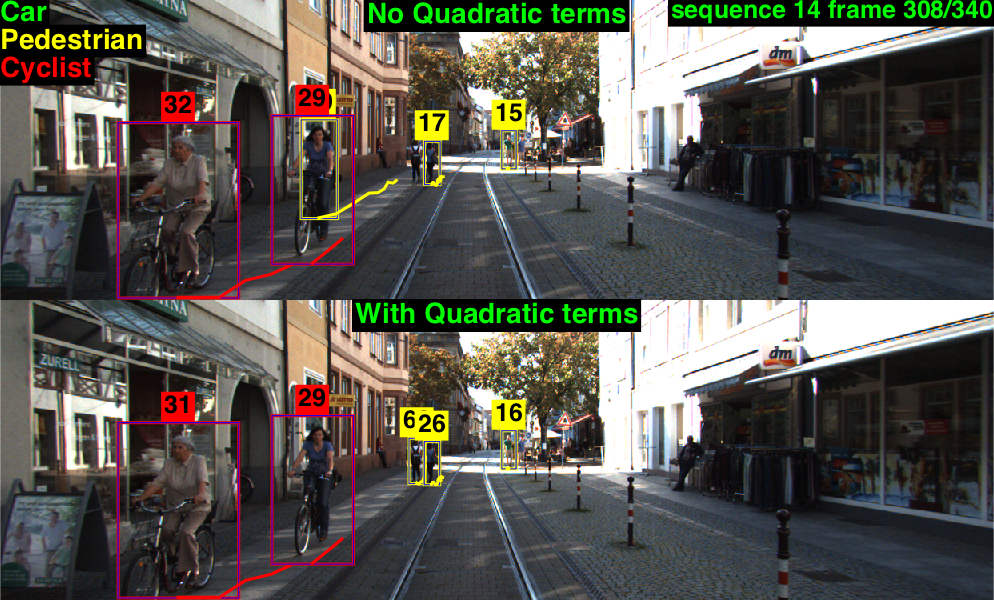}&
\includegraphics[clip,trim=0cm 0cm 0cm 0cm,width=0.47\textwidth,height=0.27\textwidth]{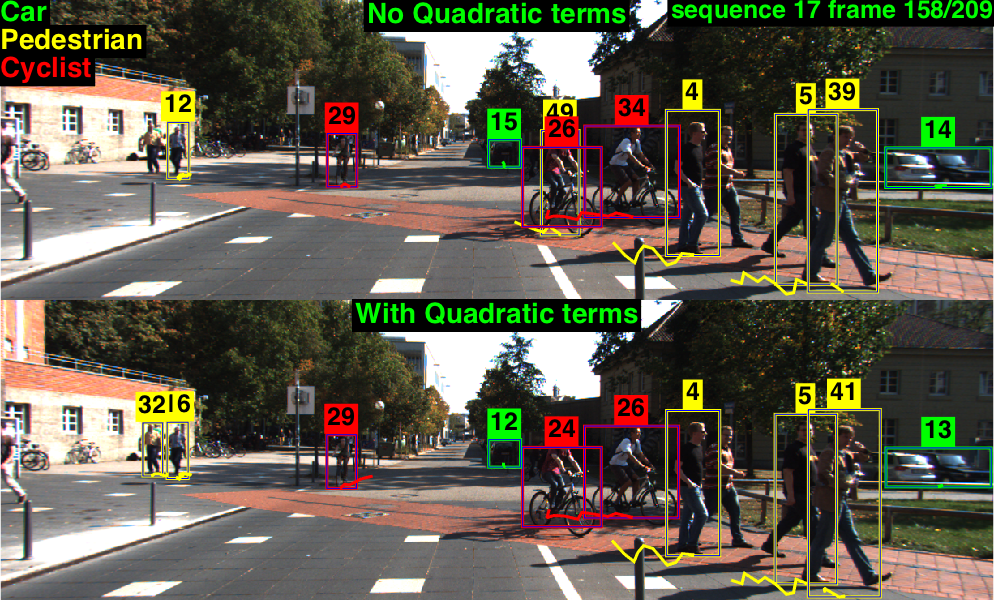}
\end{tabular}
\end{center}
\caption{Our tracking framework incorporates quadratic interactions between
objects in order to resolve appearance ambiguity and to boost weak detections.
The parameters of the interactions are learned from training examples, allow 
the tracker to successfully learn mutual exclusion between cyclist and
pedestrian, and boost to intra-class co-occurrence of nearby people.}
\label{fig:fig1}
\end{figure*}

Multi-target tracking is an active area of research in computer vision, 
driven in part by the desire to build autonomous systems that can navigate
in crowded urban environments (see e.g., \cite{Geiger2013IJRR}).
Thanks to advances of object detector performance in single, static images,
``tracking-by-detection'' approaches that build tracks on top of a collection
of candidate object detections have shown great promise. Tracking-by-detection
avoids some problems such as drift and is often able to recover from extended
periods of occlusion since it is ``self-initializing''.  Finding an optimal set
of detections corresponding to each track can often be formulated as a discrete
optimization problem of selecting a set of low-cost paths through a graph of
candidate detections for which there are efficient combinatorial algorithms
(such as min-cost matching or min-cost network-flow) that yield globally
optimal solutions (\cite{NetworkFlow,PirsiavashRF_CVPR_2011}).

Tracking by detection differs from traditional generative formulations of
multi-target tracking, which draw a distinction between the problem of
estimating a latent continuous trajectory for each object from the discrete
per-frame data association problem of assigning observations (\eg, detections)
to underlying tracks. Such methods
(e.g., \cite{Andriyenko:2012:DCO,Milan:2013:DTE,WuThScBe2012}) allow for explicitly
specifying an intuitive model of trajectory smoothness but face a difficult
joint inference problem over both continuous and discrete variables which 
can seldom be solved with any guarantee of optimality.

In tracking by detection, trajectories are implicitly defined by the selected
group of detections associated with a track.  For example, a track may skip
over some frames entirely due to occlusions or missing detections.  The
transition cost of utilizing a given edge between detections in successive
frames thus could be interpreted as some approximation of the marginal
likelihood associated with integrating over a set of underlying continuous
trajectories associated with the corresponding pair of detections.  This
viewpoint immediately raises difficulties, both in (1) encoding strong
trajectory models with only pairwise potentials and (2) identifying the
parameters of these potentials from training data.

The contribution of this paper is in demonstrating that carefully optimizing
the parameters of relatively simple combinatorial tracking-by-detection models
can yield state-of-the-art performance on difficult tracking benchmarks. 
Building on our preliminary work (\cite{WangF_BMVC_2015}), we introduce a
simple multi-target, multi-category tracking model that extends min-cost flow
with quadratic interactions between tracks in order to capture contextual
interactions within a frame.  To perform inference, we propose a family of
greedy-dynamic programming algorithms that produce high-quality solutions on
par with linear programming relaxations of the quadratic tracking objective
while being substantially faster than a general purpose LP solver.  

For learning, we use a structured prediction SVM~(\cite{Taskar03M3N}) to
optimize the complete set of tracking parameters from labeled training data. 
Structured prediction has been applied in tracking to learning inter-frame
affinity metrics~(\cite{Kim:2012:OMT:2482048.2482058}) and
association~(\cite{lou_11_structured}) as well as a variety of other learning
tasks such as fitting CRF parameters for segmentation~(\cite{Szummer_2008_ECCV}) and
word alignment for machine
translation~(\cite{Lacoste-Julien:2006:WAV:1220835.1220850}).

The structure of the remainder of the paper is as follows.  We provide a brief
overview of recent related work in Section \ref{sec:related_works} and review
the now classical network-flow model for multi-target tracking in Section
\ref{sec:model} before introducing our quadratic interaction model.  In Section
\ref{sec:inference} we describe inference algorithms for network-flow models
with quadratic interactions, namely, a standard LP-relaxation and rounding
method, and a family of novel greedy dynamic programming algorithms that can
handle quadratic interactions.  In Section \ref{sec:features} we describe the
features we used for learning tracking potentials in network-flow model with
quadratic costs.  In Section \ref{sec:learning} we describe an approach to
joint learning of model parameters in order to maximize tracking performance on
a training data set using techniques for structured prediction. We conclude
with experimental results (Section \ref{sec:experiments}) which demonstrate
that with properly learned parameters, even the basic network-flow yields
better results than many state-of-the-art methods on challenging MOT and KITTI
benchmarks. We also find that quadratic terms offer further improvements in
performance for multi-category object tracking.

\section{Related Work}
\label{sec:related_works}
Multi-target tracking problems have been tackled in a number of different 
ways.  One approach is to first group detections into candidate tracklets and
then perform scoring and association of these
tracklets~(\cite{Yang12anonline,Brendel11multiobjecttracking,Wang_2014_CVPR}).
Compared to individual detections, tracklets allow for evaluating much richer
trajectory and appearance models while maintaining some benefits of purely
combinatorial grouping. However since the problem is at least two-layered
(tracklet-generation and tracklet-association), these models are difficult to
reason about mathematically and typically lack guarantees of optimality.
Furthermore, tracklet-generation can only be done offline (or with substantial
latency) and thus approaches that rely on tracklets are inherently limited in
circumstances where online tracking is desired.

An alternative to tracklets is to attempt to include higher-order constraints
directly in a combinatorial
framework~(\cite{Butt_2013_ICCV_Workshops,ChariCVPR15}). Such methods often
operate directly over raw detections, either in an online or offline-fashion.
Offline formulations benefit from having a single well-defined objective
function or likelihood, and thus can give either globally optimal solution
(\cite{NetworkFlow}), or provide approximate solution with a certificate of
(sub)optimality (\cite{ChariCVPR15}). \cite{TangCVPR2015} propose a subgraph
multi-cut approach which differs from traditional ``path-finding'' algorithms
such as~\cite{NetworkFlow},~\cite{PirsiavashRF_CVPR_2011}
and~\cite{Butt_2013_ICCV_Workshops}.  Although designed to work directly on raw
detections, in practice~\cite{TangCVPR2015} use tracklets to reduce the
dimension of the inference problem.  Such is the trade-off between finding
globally optimal solutions and using rich tracking features.

\cite{Andriyenko:2012:DCO} attempt to solve both data association and
trajectory smoothing problem simultaneously, which results in a problem with
varying dimensionality and difficult approximate inference.~\cite{BrauICCV2013}
avoid this varying dimensionality problem by integrating out
trajectory-related variables and using Markov Chain Monte Carlo sampling to
estimate the marginal likelihoods for data association and trajectory
estimation.~\cite{SegalICCV2013} propose yet another way to avoid varying
dimensionality: instead of explicitly enumerating number of tracks, they assign
a latent variable for each real detection/track and conduct data association on these
latent variables.

Online tracking algorithms take advantage of previously identified track
associations to build rich feature models over past trajectories that
facilitate data association at the current frame.  The capability to perform
streaming data association on incoming video-frames without seeing the entire
video is a desirable property for real-time applications such as autonomous
driving. However, when the whole video is available, online tracking may make
errors that are avoidable in offline algorithms that access future frames to
resolve ambiguities.  ~\cite{KimICCV15} revisit the legacy Multiple
Hypothesis Tracking method and introduc a novel online recursive appearance
filter.~\cite{ChoiICCV15} proposes a novel flow-descriptor designed specifically
for multi-target tracking and introduces a delay period to allow correction of
possible errors made in previous frames (thus the name ``near online''), which
yields state-of-the-art accuracy.  ~\cite{SoleraICCV2015} use the relatively
simple Hungarian Matching with the novel extension to choose either ``simple''
or ``complex'' features depending on the difficulty of the inference problem at
each frame.

For any multi-target tracking approach, there are a large number of associated
model parameters which must be accurately tuned to achieve high performance.
This is particularly true for (undirected) combinatorial models based on, \eg,
network-flow, where parameters have often been set empirically by hand or
learned using piecewise training.  \cite{SoleraICCV2015}
and~\cite{DehghanCVPR2015} both use structured SVM to learn parameters of their
online data association models.  \cite{choi_eccv12} use structured SVM to learn
parameters for offline multi-target tracking with quadratic interactions for
the purpose of activity recognition. Our work differs in that it focuses on
generic activity-independent tracking and global end-to-end formulation of the
learning problem.  In particular, we develop a novel loss function that
penalizes false transition and id-errors based on the MOTA
(\cite{Bernardin:2008:EMO:1384968.1453688}) tracking score.

Finally, recent work has also pursued detectors which are specifically
optimized for tracking scenarios.  ~\cite{TangICCV2013} propose to learn
multi-person detector by using hard-negatives acquired from a tracker's output,
in order to let the detector to solve ambiguities that the tracker cannot
handle.~\cite{DehghanCVPR2015} propose to use a target identity-aware
network-flow model to process videos in batches of frames, and learn people
detectors for each individual person in an online fashion.

\section{Models for Multi-target Data Association}
\label{sec:model}
We begin by formulating multi-target tracking and data association as a
min-cost network flow problem equivalent to that of 
\cite{NetworkFlow}, where individual tracks are described by a
first-order Markov Model whose state space is a set of spatial-temporal
locations in a video.  This framework incorporates a state transition likelihood that
generates dynamics associated with a pair of successive detections along a
track, and an observation likelihood that generates appearance features for
objects and background in a given frame.  In the subsequent section we augment
this model with quadratic interactions between pairs of tracks.

\subsection{Tracking by Min-cost Flow}
For a given video sequence, we consider a discrete set of candidate object
detection sites $V$ where each candidate site $x=(l,\sigma,t) \in V$ is
a tuple described by its location $l$, scale $\sigma$ and discrete time $t$.  We
write $\Phi = \{\phi_a(x) | x \in V\}$ for the appearance features 
(image evidence) extracted at each corresponding spatial-temporal location in a video.
A single tracked object consists of an ordered set of detection sites, $T =
\{x_1, ...  , x_n\}$, where the times of successive sites are strictly
increasing.  

We model the whole video by a collection of tracks $\mathcal{T} = \{T_1, ...  ,
T_k\}$, each of which independently generates foreground object appearances at
the corresponding sites according to distribution $p_{fg}(\phi_a)$ while the
remaining site appearances are generated by a background distribution
$p_{bg}(\phi_a)$.  Each site can only belong to at most a single track which
we express by the constraint $\mathcal{T} \in \Omega$. We use
$\mathcal{B} = V \setminus \bigcup\limits_{T \in \mathcal{T}} T$ to denote the sites
which are unclaimed by any track.  Our task is to infer a collection of tracks
that maximize the posterior probability $P(\mathcal{T}|\Phi)$.  Assuming that
tracks behave independently of each other and follow a first-order Markov
model, we can write an expression for MAP inference:
\begin{align}
\mathcal{T}^* &= \underset{\mathcal{T} \in \Omega}{\operatorname{argmax}}  \; P(\Phi|\mathcal{T}) \times P(\mathcal{T}) \nonumber \\
\begin{split}
  &= \underset{\mathcal{T} \in \Omega}{\operatorname{argmax}} \; \prod_{T \in \mathcal{T}} \big[ \prod_{x \in T} p_{fg}(\phi_a(x)) \big] \prod_{x \in \mathcal{B}} p_{bg}(\phi_a(x)) \times \\
& \quad \quad \quad \quad \prod_{T \in \mathcal{T}} \big[ p_s(x_1) \prod_{i=1}^{n-1} p_t(x_{i+1}|x_i) p_e(x_n)  \big]
\end{split} 
\end{align}
where $p_s$, $p_e$ and $p_t$ represent the likelihoods for tracks starting, ending
and transitioning between given sites. Dividing through $\prod_{x \in V} p_{bg}(\phi_a(x))$
yields an equivalent problem that depends only on the appearance features at active
track locations:
\begin{align}
\begin{split}
\mathcal{T}^* &= \underset{\mathcal{T} \in \Omega}{\operatorname{argmax}}  \prod_{T \in \mathcal{T}} \big[ \prod_{x \in T} l(\phi_a(x)) \big]  \times \\
& \quad \quad \prod_{T \in \mathcal{T}} \big[ p_s(x_1) \prod_{i=1}^{n-1} p_t(x_{i+1}|x_i) p_e(x_n)  \big]
\end{split}\label{eqn:maptrack}
\end{align}
where
\begin{align}
l(\phi_a(x)) = \frac{p_{fg}(\phi_a(x))}{p_{bg}(\phi_a(x))} \nonumber
\end{align}
is the appearance likelihood ratio that a specific location $x$ corresponds to
the object tracked. 

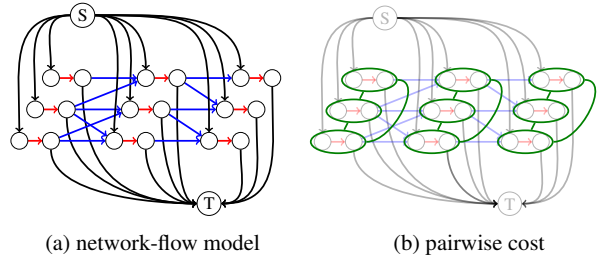
\begin{figure}[t]
\begin{center}
\begin{subfigure}[b]{0.225\textwidth}
\resizebox{0.95\textwidth}{!}{
\begin{tikzpicture}
	\begin{pgfonlayer}{nodelayer}
		\node [style=vertex_default] (0) at (-7, 1) {};
		\node [style=vertex_default] (1) at (-7.5, 0) {};
		\node [style=vertex_default] (2) at (-8, -1) {};
		\node [style=vertex_default] (3) at (-6, 1) {};
		\node [style=vertex_default] (4) at (-6.5, 0) {};
		\node [style=vertex_default] (5) at (-7, -1) {};
		\node [style=vertex_default] (6) at (-4, 1) {};
		\node [style=vertex_default] (7) at (-4.5, 0) {};
		\node [style=vertex_default] (8) at (-3, 1) {};
		\node [style=vertex_default] (9) at (-3.5, 0) {};
		\node [style=vertex_default] (10) at (-1, 1) {};
		\node [style=vertex_default] (11) at (-1.5, 0) {};
		\node [style=vertex_default] (12) at (0, 1) {};
		\node [style=vertex_default] (13) at (-0.5, 0) {};
		\node [style=vertex_default] (14) at (-6, 3) {\LARGE{S}};
		\node [style=vertex_default] (15) at (-2, -3) {\LARGE{T}};
		\node [style=vertex_default] (16) at (-1, -1) {};
		\node [style=vertex_default] (17) at (-2, -1) {};
		\node [style=vertex_default] (18) at (-4, -1) {};
		\node [style=vertex_default] (19) at (-5, -1) {};
	\end{pgfonlayer}
	\begin{pgfonlayer}{edgelayer}
		\draw [style=edge_forward,draw=red] (0) to (3);
		\draw [style=edge_forward,draw=red] (1) to (4);
		\draw [style=edge_forward,draw=red] (2) to (5);
		\draw [style=edge_forward,draw=red] (6) to (8);
		\draw [style=edge_forward,draw=red] (7) to (9);
		\draw [style=edge_forward,draw=red] (10) to (12);
		\draw [style=edge_forward,draw=red] (11) to (13);
		\draw [style=edge_forward, in=90, out=180, looseness=1.25] (14) to (0);
		\draw [style=edge_forward, in=90, out=180, looseness=1.25] (14) to (1);
		\draw [style=edge_forward, in=90, out=180, looseness=1.25] (14) to (2);
		\draw [style=edge_forward, in=90, out=0, looseness=1.25] (14) to (6);
		\draw [style=edge_forward, in=90, out=0, looseness=1.25] (14) to (7);
		\draw [style=edge_forward, in=90, out=0] (14) to (10);
		\draw [style=edge_forward, in=90, out=0, looseness=1.50] (14) to (11);
		\draw [style=edge_forward, in=180, out=-90, looseness=0.75] (5) to (15);
		\draw [style=edge_forward, in=180, out=-90] (4) to (15);
		\draw [style=edge_forward, in=180, out=270, looseness=1.25] (3) to (15);
		\draw [style=edge_forward, in=180, out=-90, looseness=0.75] (9) to (15);
		\draw [style=edge_forward, in=180, out=-90, looseness=0.50] (8) to (15);
		\draw [style=edge_forward, in=0, out=-90, looseness=1.25] (13) to (15);
		\draw [style=edge_forward, in=0, out=-90, looseness=1.25] (12) to (15);
		\draw [style=edge_forward,draw=blue] (3) to (6);
		\draw [style=edge_forward,draw=blue] (4) to (7);
		\draw [style=edge_forward,draw=blue] (5) to (7);
		\draw [style=edge_forward,draw=blue] (4) to (6);
		\draw [style=edge_forward,draw=blue] (8) to (10);
		\draw [style=edge_forward,draw=blue] (9) to (11);
		\draw [style=edge_forward,draw=blue] (8) to (11);
		\draw [style=edge_forward,draw=red] (17) to (16);
		\draw [style=edge_forward,draw=red] (19) to (18);
		\draw [style=edge_forward,draw=blue] (5) to (19);
		\draw [style=edge_forward,draw=blue] (18) to (17);
		\draw [style=edge_forward, in=90, out=0, looseness=0.75] (14) to (19);
		\draw [style=edge_forward, in=180, out=-90] (18) to (15);
		\draw [style=edge_forward, in=90, out=0, looseness=1.75] (14) to (17);
		\draw [style=edge_forward, in=0, out=-90, looseness=1.25] (16) to (15);
		\draw [style=edge_forward,draw=blue] (4) to (19);
		\draw [style=edge_forward,draw=blue] (9) to (17);
	\end{pgfonlayer}
\end{tikzpicture}
}
\caption{network-flow model}
\end{subfigure}
\begin{subfigure}[b]{0.245\textwidth}
\resizebox{0.95\textwidth}{!}{
\begin{tikzpicture}
	\begin{pgfonlayer}{nodelayer}
		\node [style=vertex_default, draw opacity=0.3, fill opacity=0.3] (0) at (-7, 1) {};
		\node [style=vertex_default, draw opacity=0.3, fill opacity=0.3] (1) at (-7.5, 0) {};
		\node [style=vertex_default, draw opacity=0.3, fill opacity=0.3] (2) at (-8, -1) {};
		\node [style=vertex_default, draw opacity=0.3, fill opacity=0.3] (3) at (-6, 1) {};
		\node [style=vertex_default, draw opacity=0.3, fill opacity=0.3] (4) at (-6.5, 0) {};
		\node [style=vertex_default, draw opacity=0.3, fill opacity=0.3] (5) at (-7, -1) {};
		\node [style=vertex_default, draw opacity=0.3, fill opacity=0.3] (6) at (-4, 1) {};
		\node [style=vertex_default, draw opacity=0.3, fill opacity=0.3] (7) at (-4.5, 0) {};
		\node [style=vertex_default, draw opacity=0.3, fill opacity=0.3] (8) at (-3, 1) {};
		\node [style=vertex_default, draw opacity=0.3, fill opacity=0.3] (9) at (-3.5, 0) {};
		\node [style=vertex_default, draw opacity=0.3, fill opacity=0.3] (10) at (-1, 1) {};
		\node [style=vertex_default, draw opacity=0.3, fill opacity=0.3] (11) at (-1.5, 0) {};
		\node [style=vertex_default, draw opacity=0.3, fill opacity=0.3] (12) at (0, 1) {};
		\node [style=vertex_default, draw opacity=0.3, fill opacity=0.3] (13) at (-0.5, 0) {};
		\node [style=vertex_default, draw opacity=0.3, fill opacity=0.3] (14) at (-6, 3) {\LARGE{S}};
		\node [style=vertex_default, draw opacity=0.3, fill opacity=0.3] (15) at (-2, -3) {\LARGE{T}};
		\node [style=vertex_default, draw opacity=0.3, fill opacity=0.3] (16) at (-1, -1) {};
		\node [style=vertex_default, draw opacity=0.3, fill opacity=0.3] (17) at (-2, -1) {};
		\node [style=vertex_default, draw opacity=0.3, fill opacity=0.3] (18) at (-4, -1) {};
		\node [style=vertex_default, draw opacity=0.3, fill opacity=0.3] (19) at (-5, -1) {};
		\node [style=vertex_hyper, fill opacity=0] (20) at (-6.5, 1){1};
		\node [style=vertex_hyper, fill opacity=0] (21) at (-7, 0){1};
		\node [style=vertex_hyper, fill opacity=0] (22) at (-7.5, -1){1};
		
		\node [style=vertex_hyper, fill opacity=0] (23) at (-3.5, 1){1};
		\node [style=vertex_hyper, fill opacity=0] (24) at (-4, 0){1};
		\node [style=vertex_hyper, fill opacity=0] (25) at (-4.5, -1){1};
		
		\node [style=vertex_hyper, fill opacity=0] (26) at (-0.5, 1){1};
		\node [style=vertex_hyper, fill opacity=0] (27) at (-1, 0){1};
		\node [style=vertex_hyper, fill opacity=0] (28) at (-1.5, -1){1};	
	\end{pgfonlayer}
	\begin{pgfonlayer}{edgelayer}
		\draw [style=edge_forward,draw=red,draw opacity=0.3] (0) to (3);
		\draw [style=edge_forward,draw=red,draw opacity=0.3] (1) to (4);
		\draw [style=edge_forward,draw=red,draw opacity=0.3] (2) to (5);
		\draw [style=edge_forward,draw=red,draw opacity=0.3] (6) to (8);
		\draw [style=edge_forward,draw=red,draw opacity=0.3] (7) to (9);
		\draw [style=edge_forward,draw=red,draw opacity=0.3] (10) to (12);
		\draw [style=edge_forward,draw=red,draw opacity=0.3] (11) to (13);
		\draw [style=edge_forward, in=90, out=180, looseness=1.25, draw opacity=0.3] (14) to (0);
		\draw [style=edge_forward, in=90, out=180, looseness=1.25, draw opacity=0.3] (14) to (1);
		\draw [style=edge_forward, in=90, out=180, looseness=1.25, draw opacity=0.3] (14) to (2);
		\draw [style=edge_forward, in=90, out=0, looseness=1.25, draw opacity=0.3] (14) to (6);
		\draw [style=edge_forward, in=90, out=0, looseness=1.25, draw opacity=0.3] (14) to (7);
		\draw [style=edge_forward, in=90, out=0, draw opacity=0.3] (14) to (10);
		\draw [style=edge_forward, in=90, out=0, looseness=1.50, draw opacity=0.3] (14) to (11);
		\draw [style=edge_forward, in=180, out=-90, looseness=0.75, draw opacity=0.3] (5) to (15);
		\draw [style=edge_forward, in=180, out=-90, draw opacity=0.5, draw opacity=0.3] (4) to (15);
		\draw [style=edge_forward, in=180, out=270, looseness=1.25, draw opacity=0.3] (3) to (15);
		\draw [style=edge_forward, in=180, out=-90, looseness=0.75, draw opacity=0.3] (9) to (15);
		\draw [style=edge_forward, in=180, out=-90, looseness=0.50, draw opacity=0.3] (8) to (15);
		\draw [style=edge_forward, in=0, out=-90, looseness=1.25, draw opacity=0.3] (13) to (15);
		\draw [style=edge_forward, in=0, out=-90, looseness=1.25, draw opacity=0.3] (12) to (15);
		\draw [style=edge_forward,draw=blue,draw opacity=0.3] (3) to (6);
		\draw [style=edge_forward,draw=blue,draw opacity=0.3] (4) to (7);
		\draw [style=edge_forward,draw=blue,draw opacity=0.3] (5) to (7);
		\draw [style=edge_forward,draw=blue,draw opacity=0.3] (4) to (6);
		\draw [style=edge_forward,draw=blue,draw opacity=0.3] (8) to (10);
		\draw [style=edge_forward,draw=blue,draw opacity=0.3] (9) to (11);
		\draw [style=edge_forward,draw=blue,draw opacity=0.3] (8) to (11);
		\draw [style=edge_forward,draw=red,draw opacity=0.3] (17) to (16);
		\draw [style=edge_forward,draw=red,draw opacity=0.3] (19) to (18);
		\draw [style=edge_forward,draw=blue,draw opacity=0.3] (5) to (19);
		\draw [style=edge_forward,draw=blue,draw opacity=0.3] (18) to (17);
		\draw [style=edge_forward, in=90, out=0, looseness=0.75, draw opacity=0.3] (14) to (19);
		\draw [style=edge_forward, in=180, out=-90, draw opacity=0.5] (18) to (15);
		\draw [style=edge_forward, in=90, out=0, looseness=1.75, draw opacity=0.3] (14) to (17);
		\draw [style=edge_forward, in=0, out=-90, looseness=1.25, draw opacity=0.3] (16) to (15);
		\draw [style=edge_forward,draw=blue,draw opacity=0.3] (4) to (19);
		\draw [style=edge_forward,draw=blue,draw opacity=0.3] (9) to (17);
		
		\draw [ultra thick,draw=Green] (20) to (21);
		\draw [ultra thick,draw=Green] (21) to (22);
		\draw [ultra thick,in=0, out=0, looseness=1,draw=Green] (20) to (22);
		
		\draw [ultra thick,draw=Green] (23) to (24);
		\draw [ultra thick,draw=Green] (24) to (25);
		\draw [ultra thick,in=0, out=0, looseness=1,draw=Green] (23) to (25);
		
		\draw [ultra thick,draw=Green] (26) to (27);
		\draw [ultra thick,draw=Green] (27) to (28);
		\draw [ultra thick,in=0, out=0, looseness=1,draw=Green] (26) to (28);
	\end{pgfonlayer}
\end{tikzpicture}
}
\caption{pairwise cost}
\end{subfigure}
\end{center}
\caption{Graphical representation of network flow model and its extension with pairwise costs.
(a) Basic network-flow model for multi-target tracking, as described by~\cite{NetworkFlow}.
A pair of nodes (connected by red edge) represent a detection, blue edges represent possible
transitions between detections and track birth/death events are modeled by black edges. Costs $c_i$ 
in Eq.~\ref{eqn:mincostflow} are for red edges, $c_{ij}$ are for blue edges, $c_i^t$ and $c_i^s$
are for black edges. (b) Network-flow model extended with pairwise cost. A green node represents
a detection site while an undirected edge between green nodes encodes the pairwise cost for choosing 
both detection sites as part of the solution. To simplify our description, in later text we will refer to a green
node (a red edge and the two nodes associated with it) as a ``detection node''. The set $V$ consists of
all detection nodes in the graph, whereas the set $E$ consists of all transition edges
in the graph.}
\label{fig:mincostflow}
\end{figure}

The set of optimal (most probable) tracks under this model can be found by
solving an integer linear program (ILP) over flow variables $\mathbf{f}$ that
indicate which detections are active in each frame $\{f_i\}$ and which pairs
of detections are associated between frames $\{f_{ij}\}$.  Figure
\ref{fig:mincostflow} shows a graphical representation where an individual
object track corresponds to a directed st-path traversing edges that encode
start, detection, transition and end costs.  By taking a negative log of the
MAP objective, this equivalent formulation can be written as:
\begin{align}
\underset{\mathbf{f}}{\operatorname{min}} &\ \sum_{i \in V} c_i^s f_i^s + \sum_{ij \in E} c_{ij} f_{ij} + \sum_{i \in V} c_i f_i + \sum_{i \in V} c_i^t f_i^t 
\label{eqn:mincostflow} \\
\text{s.t.}&\quad f_i^s + \sum_j f_{ji} = f_i = f_i^t + \sum_j f_{ij} \label{eqn:flowconstraint}\\
& f_i^s, f_i^t, f_i, f_{ij} \in \{0,1\} \label{eqn:integerconstraint1}
\end{align}
where $E$ is the set of valid transitions between sites in successive 
frames and the costs are given by:
\begin{align}
\begin{split}
c_i = -\log \frac{p_{fg}(\phi_a(x_i))}{p_{bg}(\phi_a(x_i))} &, \quad c_{ij} = -\log p_t(x_j | x_i),\\
c_i^s = -\log p_s(x_i) &, \quad c_i^t = -\log p_e(x_i) 
\label{eqn:potentials}
\end{split}
\end{align}
and the integrality constraint on $\mathbf{f}$ enforces the requirement
that each site belongs to at most a single track.

This ILP is a well studied problem known as minimum-cost network
flow~(\cite{SSP}) with unit capacity edges.  In particular, the flow
constraints satisfy the \textit{total unimodularity} property and thus an
integral solution can be found by LP relaxation or via efficient
specialized solvers such as network simplex, successive shortest path and
push-relabel with bisectional search~(\cite{NetworkFlow}).

\subsection{Track Interdependence}
The aforementioned model assumes tracks are independent of each other, which is
not always true in practice.  In order to allow interactions between multiple objects,
we add a pairwise cost term denoted $q_{ij}$ for jointly
activating a pair of flows $f_i$ and $f_j$ corresponding to detections at 
sites $x_i = (l_i,\sigma_i,t_i)$ and $x_j=(l_j,\sigma_j,t_j)$. Adding this
term to Eq.~\ref{eqn:mincostflow} yields an Integer Quadratic Program
(IQP):
\begin{align}
\begin{split}
\underset{\mathbf{f}}{\operatorname{min}} &\ \sum_{i \in V} c_i^s f_i^s + \sum_{ij \in E} c_{ij} f_{ij} + \sum_{i \in V} c_i f_i + \sum_{i \in V} c_i^t f_i^t\\
&+ \sum_{ij \in EC } q_{ij} f_i f_j
\end{split} \label{eqn:quadflow} \\
\text{s.t.}&\quad f_i^s + \sum_j f_{ji} = f_i = f_i^t + \sum_j f_{ij} \nonumber\\
& \quad f_i^s, f_i^t, f_i, f_{ij} \in \{0,1\} \nonumber
\end{align}
In our experiments, we only consider pairwise interactions between pairs of sites in the
same video frame which we denote by $EC = \{ij : t_i = t_j\}$. One could easily
extend such formulation to include transition-transition interactions to model
high order dynamics.

Unlike min-cost flow (Eq.~\ref{eqn:mincostflow}), finding the global minimum of
the IQP problem (Eq.~\ref{eqn:quadflow}) is NP-hard~(\cite{ANasser14}) due to
the quadratic terms in the objective.  In the next section we discuss two
different approximations for finding high-quality solutions $\mathbf{f}$. In
Section~\ref{sec:learning} we describe how the costs $\mathbf{c}$ and $\mathbf{q}$
can be learned from data.

\section{Inference}
\label{sec:inference}
We evaluate two different schemes for finding high-quality approximate
solutions to the quadratic tracking objective.  The first is a standard
approach of introducing auxiliary variables and relaxing the integrality
constraints to yield a linear program (LP) that lower-bounds the original
objective. We also consider a family of greedy approximation algorithms based
on successive rounds of dynamic programming that also yields good solutions
while avoiding the expense of solving a large scale LP. The resulting tracks
(encoded by the optimal flows $\mathbf{f}$) are used for both test-time track
prediction as well as for optimizing parameters during learning (see
Section~\ref{sec:learning}).

\subsection{LP Relaxation and Rounding}
\label{sec:lp}
If we relax the integer constraints and deform the costs as necessary to make
the objective convex, then the global optimum of Eq.~\ref{eqn:quadflow} can be
found in polynomial time. For example, one could apply Frank-Wolfe algorithm
to optimize the relaxed, convexified QP while simultaneously keeping track of
good integer solutions~(\cite{TangECCV14}).  However, for real-world tracking
over long videos, the relaxed QP is still quite expensive to solve. Instead
we follow the approach proposed by
\cite{ChariCVPR15}, reformulating the IQP as an
equivalent ILP problem by replacing the quadratic terms $f_i f_j$ with a set of
auxiliary variables $u_{ij}$:
\begin{align}
\begin{split}
\underset{\mathbf{f}}{\operatorname{min}} &\ \sum_{i \in V} c_i^s f_i^s + \sum_{ij \in E} c_{ij} f_{ij} + \sum_{i \in V} c_i f_i + \sum_{i \in V} c_i^t f_i^t\\
& + \sum_{ij \in EC } q_{ij} u_{ij}
\end{split} \label{eqn:lprelax} \\
\begin{split}
\text{s.t.}&\quad f_i^s + \sum_j f_{ji} = f_i = f_i^t + \sum_j f_{ij} \nonumber\\
& \quad f_i^s, f_i^t, f_i, f_{ij} \in \{0,1\} \nonumber \\
& \quad u_{ij} \le f_i, u_{ij} \le f_j \quad f_i + f_j \le u_{ij} + 1 \nonumber
\end{split}
\end{align}
The new constraint sets enforce $u_{ij}$ to be $1$ only when $f_i$ and $f_j$
are both $1$. By relaxing the integrality constraints, the program in Eq.
\ref{eqn:lprelax} can be solved efficiently via large scale LP solvers
such as CPLEX or MOSEK.

During test time we would like to predict a discrete set of tracks. This
requires rounding the solution of the relaxed LP to some solution that
satisfies not only integer constraints but also flow constraints.
~\cite{ChariCVPR15} proposed two rounding heuristics: a
Euclidean rounding scheme that minimizes $\| \mathbf{f} - \widehat{\mathbf{f}}
\|^2$ where $\widehat{\mathbf{f}}$ is the non-integral solution given by the LP
relaxation. When $\mathbf{f}$ is constrained to be binary, this objective
simplifies to a linear function $(\mathbf{1}-2 \widehat{\mathbf{f}})^T \mathbf{f} + \|
\widehat{\mathbf{f}} \|^2$, which can be optimized using a standard linear
min-cost flow solver.  Alternately, one can use a linear under-estimator of
the objective in Eq.~\ref{eqn:quadflow}, similar to the Frank-Wolfe algorithm:
\begin{align}
\begin{split}
&\underset{\mathbf{f}}{\operatorname{min}} \sum_{i \in V} c_i^s f_i^s + \sum_{ij \in E} c_{ij} f_{ij} + \sum_{i \in V} c_i^t f_i^t \\
& \quad \quad +\sum_{i \in V} ( c_i + \sum_{ij \in EC} q_{ij} \widehat{u}_{ij} + \sum_{ji \in EC} q_{ji} \widehat{u}_{ji} ) f_i 
\end{split} 
\end{align}
Both of these rounding heuristics involve optimizing a new linear objective function
subject to the original integer and flow constraints and thus can be solved as 
an ordinary min-cost network flow problem. In our experiments we execute both
rounding heuristics and choose the solution with lower cost under the original 
quadratic objective.

\subsection{Greedy Dynamic Programming}
\label{sec:dp}
As an alternative to the LP relaxation, we describe a family of greedy algorithms
inspired by the combination of dynamic programming (DP) proposed by ~\cite{PirsiavashRF_CVPR_2011}
for approximately solving min-cost flow and the 
greedy forward selection used for modeling contextual object interactions
by~\cite{DesaiRF_ICCV_2009}. Our general strategy is to sequentially push units
of flow through the tracking graph, updating the edge costs at each step to capture
the expected contribution of quadratic interactions.

\subsubsection{Successive Shortest Paths}
\label{sec:ssp}
We start by briefly describing the successive shortest path (SSP) algorithm
which solves for the min-cost flow with the standard linear objective and refer
the reader to \cite{SSP} for a comprehensive discussion. Consider the example
tracking graph shown in Figure~\ref{fig:mincostflow}. SSP finds the global optimum of
Eq.~\ref{eqn:mincostflow} by repeatedly searching for shortest st-path in
a so-called {\em residual graph}.  In our model where all edges have unit capacity, the
residual graph $G_r(\mathbf{f})$ associated with flow $\mathbf{f}$ is given by
reversing the orientation of every directed edge used by the solution $\mathbf{f}$ 
and negating their associated costs.

Starting from an empty flow $\mathbf{f}=0$, SSP simply iterates two steps,
\begin{enumerate}
  \item Find the minimum cost st-path in the residual graph $G_r(\mathbf{f})$.
  \item If the cost of the path is negative, push a unit flow along the path and update $\mathbf{f}$.
\end{enumerate}
until no negative cost path can be found.  

We note that the first iteration of SSP looks like a single-target data association problem.
In particular, one can utilize a simple dynamic program that makes a single sweep over 
the graph nodes ordered by time to find the minimum cost (shortest) path. However, the 
residual graph in subsequent iterations contains negative weight edges and is no longer 
acyclic, hence requiring the use of a more general and computationally expensive shortest 
path algorithm such as Bellman-Ford.

Here we consider fast approximations to the full 
shortest path problem based on carrying out multiple temporally-ordered sweeps of the 
graph.  This ``K-Pass Dynamic Program'' approach introduced 
by~\cite{PirsiavashRF_CVPR_2011} can be viewed as a variant
of Bellman-Ford that only considers a subset of possible shortest paths using a 
problem-specific schedule for performing edge relaxations.  This approximate approach 
has been shown to provide solutions which are nearly as good as the full shortest path
computation and provides a natural way to incorporate quadratic cost terms.

\subsubsection{One-pass Dynamic Programming}
\label{sec:DP1}

Assume the detection nodes are sorted in time. We denote 
$cost(i)$ as the cost of the shortest path from source node to node $i$, $link(i)$ as 
$i$'s predecessor in this shortest path, and
$birth\_node(i)$ as the first detection node in this shortest path. We initialize 
$cost(i) = c_i + c_i^s$, $link(i) = \emptyset$, and $birth\_node(i) = i$ for all $i \in V$.

To find the shortest path on the initial DAG $G$, we can sweep from first frame 
to last frame, computing $cost(i)$ as:
\begin{align}
cost(i) = c_i + \underset{j : ij \in E}{\operatorname{min}} \; \left\{c_i^s, \;  c_{ij} + cost(j)\right\}
\end{align}
and store the argmin in $birth\_node(i)$ or $link(i)$ accordingly.

After sweeping through all frames, we find a node $i$ such that $cost(i) +
c_i^t$ is minimum and reconstruct the corresponding shortest path (which 
terminates at $i$ and has cost $cost(i)+c_i^t$) by backtracking cached $link$
variables.  After the shortest path is identified, we remove all nodes and
edges in this path from $G$; the resulting graph $G'$ will still be a DAG.
Without quadratic terms we can just repeat this procedure until we cannot find
any path that has a negative cost. This can be viewed as a greedy version of
SSP that does not change paths once they have been added to the solution.

\paragraph{Quadratic cost updates} The sequential greedy nature of 
one-pass dynamic programming is well suited to incorporating estimates of the quadratic 
cost terms in Eq.~\ref{eqn:quadflow} by performing an additional contextual update 
step after each round of dynamic programming.  The complete algorithm is 
outlined in Algorithm~\ref{alg:alg1}.  After each new track is instanced, the 
edge costs in the associated flow graph used for finding tracks in subsequent
iterations are updated to include the quadratic penalties or boosts incurred by 
the newly instanced track.  As was the case with the LP+rounding scheme described 
in Section~\ref{sec:lp}, Algorithm~\ref{alg:alg1} does not guarantee an optimal
solution. However, as we show in the experiments, it performs well in practice.

\begin{algorithm}{}
\begin{minipage}{0.45\textwidth}
\caption{One-pass DP with Quadratic Cost Update}
\label{alg:alg1}
\begin{algorithmic}[1]
\State \textbf{Input}: A Directed-Acyclic-Graph $G$ with edge weights $c_i,c_{ij}$
\State initialize $\mathcal{T} \leftarrow \emptyset$
\Repeat
  \State Find shortest st-path $\mathbf{p}$ on $G$ via dynamic programming
  \State $track\_cost = cost(\mathbf{p})$
  \If {$track\_cost < 0$}
    \ForAll{locations $x_i$ in $\mathbf{p}$}
      \State $c_j = c_j + q_{ij} + q_{ji}$ for all $ij \in EC$
      \State $c_i = +\infty$
    \EndFor
    \State $\mathcal{T} \leftarrow \mathcal{T} \cup \mathbf{p}$
  \EndIf
\Until{$track\_cost \ge 0$}
\State \textbf{Output}: track collection $\mathcal{T}$
\end{algorithmic}
\end{minipage}
\end{algorithm}

\subsubsection{Two-pass Dynamic Programming}
\label{sec:DP2}
Unlike a general shortest path algorithm such as Bellman-Ford, one-pass
dynamic programming can only find shortest paths consisting of forward going
edges.  As proposed by~\cite{PirsiavashRF_CVPR_2011}, one can improve this
approximation by carrying out multiple passes of DP that run forward and
backward in time to find paths in the residual graph $G_r(\mathbf{f})$ that
reverse temporal direction one or more times.  First we describe the details of
2-pass dynamic programming without quadratic contextual updates.

Let $V_f$ denote the set of forward edges in the current residual graph, \ie,
detection and transition variables in $\mathbf{f}$ that equal 0, and 
$V_b$ as the set of backward edges in current residual graph,
\ie, variables in $\mathbf{f}$ that equal 1. The 2-pass DP algorithm
proceeds as follows:
\begin{enumerate}
  \item Perform a pass of forward DP (from first frame to last frame) on all nodes.
  When computing $cost$ of a specific node, simply ignore all its predecessors that
  belong to $V_b$.
  \item Set $cost_{en}(i) = cost(i) - c_i$ for all $i \in V_b$ and 
  perform one pass of backward DP (from last frame to first frame) on $V_b$. Update
  $cost(i)$ and $cost_{en}(i)$ for $i \in V_b$ at each step,
  \begin{align}
  &cost(i) = cost_{en}(j)-c_{ij}\\
  &cost_{en}(i) = {\operatorname{min}} \;(cost_{en}(i), cost(i) - c_i)
  \end{align}
  where $j$ is $i$'s backward predecessor and $c_{ij}$ is from the original graph. Set
  $cost(i) = +\infty$ for any backward detection $i$ that has no backward
  transition edge coming to it (\ie, the last node of each instanced track).
  \item Perform a second pass of forward DP on $i \in V_f$. To avoid running into cyclic path,
  we need to backtrack shortest paths for all $j \in N(i)$, where $N(i)$ is all neighboring
  nodes that are connected to $i$ via a forward edge. In practice we only need to check
  $j \in N(i)$ such that $j$ and $i$ share the same birth node, as they cannot
  form a cyclic path otherwise. Furthermore, one can keep a cache of shortest path so far
  for each node, reducing the backtrack to a constant $O(1)$ operation.
  \item Find node $i$ with minimum $cost(i) + c_i^t$, the (approximate) shortest path is
  then $path(i)$.
  \item Update solution $\mathbf{f}$ by setting all forward variables along
  $path(i)$ to 1 and all backward variables along $path(i)$ to 0.
\end{enumerate}

It is straightforward to see that during the first iteration, 1-pass DP and
2-pass DP behave identically. Since we enforce the path found by 2-pass DP
never goes into a source node or out of a sink node, each iteration generates
exactly one more track (either by splitting a previously found track into two
or by instancing an entirely new track). Therefore the algorithm will terminate
after at most $|V|$ iterations.

\begin{figure}[t]
\begin{center}
\begin{subfigure}[b]{0.225\textwidth}
\resizebox{0.95\textwidth}{!}{
\begin{tikzpicture}
	\begin{pgfonlayer}{nodelayer}
		\node [style=vertex_default] (0) at (-7, 1) {};
		\node [style=vertex_default] (1) at (-7.5, 0) {};
		\node [style=vertex_default] (2) at (-8, -1) {};
		\node [style=vertex_default] (3) at (-6, 1) {};
		\node [style=vertex_default] (4) at (-6.5, 0) {};
		\node [style=vertex_default] (5) at (-7, -1) {};
		\node [style=vertex_default] (6) at (-4, 1) {};
		\node [style=vertex_default] (7) at (-4.5, 0) {};
		\node [style=vertex_default] (8) at (-3, 1) {};
		\node [style=vertex_default] (9) at (-3.5, 0) {};
		\node [style=vertex_default] (10) at (-1, 1) {};
		\node [style=vertex_default] (11) at (-1.5, 0) {};
		\node [style=vertex_default] (12) at (0, 1) {};
		\node [style=vertex_default] (13) at (-0.5, 0) {};
		\node [style=vertex_default] (14) at (-6, 3) {\LARGE{S}};
		\node [style=vertex_default] (15) at (-2, -3) {\LARGE{T}};
		\node [style=vertex_default] (16) at (-1, -1) {};
		\node [style=vertex_default] (17) at (-2, -1) {};
		\node [style=vertex_default] (18) at (-4, -1) {};
		\node [style=vertex_default] (19) at (-5, -1) {};
	\end{pgfonlayer}
	\begin{pgfonlayer}{edgelayer}
		\draw [style=edge_forward] (0) to (3);
		\draw [style=edge_forward] (1) to (4);
		\draw [style=edge_forward] (2) to (5);
		\draw [style=edge_forward] (6) to (8);
		\draw [style=edge_forward] (7) to (9);
		\draw [style=edge_forward] (10) to (12);
		\draw [style=edge_forward] (11) to (13);
		\draw [style=edge_forward, in=90, out=180, looseness=1.25] (14) to (0);
		\draw [style=edge_forward, in=90, out=180, looseness=1.25] (14) to (1);
		\draw [style=edge_forward, in=90, out=180, looseness=1.25] (14) to (2);
		\draw [style=edge_forward, in=90, out=0, looseness=1.25] (14) to (6);
		\draw [style=edge_forward, in=90, out=0, looseness=1.25] (14) to (7);
		\draw [style=edge_forward, in=90, out=0] (14) to (10);
		\draw [style=edge_forward, in=90, out=0, looseness=1.50] (14) to (11);
		\draw [style=edge_forward, in=180, out=-90, looseness=0.75] (5) to (15);
		\draw [style=edge_forward, in=180, out=-90] (4) to (15);
		\draw [style=edge_forward, in=180, out=270, looseness=1.25] (3) to (15);
		\draw [style=edge_forward, in=180, out=-90, looseness=0.75] (9) to (15);
		\draw [style=edge_forward, in=180, out=-90, looseness=0.50] (8) to (15);
		\draw [style=edge_forward, in=0, out=-90, looseness=1.25] (13) to (15);
		\draw [style=edge_forward, in=0, out=-90, looseness=1.25] (12) to (15);
		\draw [style=edge_forward] (3) to (6);
		\draw [style=edge_forward] (4) to (7);
		\draw [style=edge_forward] (5) to (7);
		\draw [style=edge_forward] (4) to (6);
		\draw [style=edge_forward] (8) to (10);
		\draw [style=edge_forward] (9) to (11);
		\draw [style=edge_forward] (8) to (11);
		\draw [style=edge_forward] (17) to (16);
		\draw [style=edge_forward] (19) to (18);
		\draw [style=edge_forward] (5) to (19);
		\draw [style=edge_forward] (18) to (17);
		\draw [style=edge_forward, in=90, out=0, looseness=0.75] (14) to (19);
		\draw [style=edge_forward, in=180, out=-90] (18) to (15);
		\draw [style=edge_forward, in=90, out=0, looseness=1.75] (14) to (17);
		\draw [style=edge_forward, in=0, out=-90, looseness=1.25] (16) to (15);
		\draw [style=edge_forward] (4) to (19);
		\draw [style=edge_forward] (9) to (17);
	\end{pgfonlayer}
\end{tikzpicture}
}
\caption{}
\end{subfigure}
\begin{subfigure}[b]{0.225\textwidth}
\resizebox{0.95\textwidth}{!}{
\begin{tikzpicture}
	\begin{pgfonlayer}{nodelayer}
		\node [style=vertex_default] (0) at (-7, 1) {};
		\node [style=vertex_default] (1) at (-7.5, 0) {};
		\node [style=vertex_default] (2) at (-8, -1) {};
		\node [style=vertex_default] (3) at (-6, 1) {};
		\node [style=vertex_default] (4) at (-6.5, 0) {};
		\node [style=vertex_default] (5) at (-7, -1) {};
		\node [style=vertex_default] (6) at (-4, 1) {};
		\node [style=vertex_default] (7) at (-4.5, 0) {};
		\node [style=vertex_default] (8) at (-3, 1) {};
		\node [style=vertex_default] (9) at (-3.5, 0) {};
		\node [style=vertex_default] (10) at (-1, 1) {};
		\node [style=vertex_default] (11) at (-1.5, 0) {};
		\node [style=vertex_default] (12) at (0, 1) {};
		\node [style=vertex_default] (13) at (-0.5, 0) {};
		\node [style=vertex_default] (14) at (-6, 3) {\LARGE{S}};
		\node [style=vertex_default] (15) at (-2, -3) {\LARGE{T}};
		\node [style=vertex_default] (16) at (-1, -1) {};
		\node [style=vertex_default] (17) at (-2, -1) {};
		\node [style=vertex_default] (18) at (-4, -1) {};
		\node [style=vertex_default] (19) at (-5, -1) {};
	\end{pgfonlayer}
	\begin{pgfonlayer}{edgelayer}
		\draw [style=edge_forward] (0) to (3);
		\draw [style=edge_forward_selected] (1) to (4);
		\draw [style=edge_forward] (2) to (5);
		\draw [style=edge_forward] (6) to (8);
		\draw [style=edge_forward_selected] (7) to (9);
		\draw [style=edge_forward] (10) to (12);
		\draw [style=edge_forward_selected] (11) to (13);
		\draw [style=edge_forward, in=90, out=180, looseness=1.25] (14) to (0);
		\draw [style=edge_forward_selected, in=90, out=180, looseness=1.25] (14) to (1);
		\draw [style=edge_forward, in=90, out=180, looseness=1.25] (14) to (2);
		\draw [style=edge_forward, in=90, out=0, looseness=1.25] (14) to (6);
		\draw [style=edge_forward, in=90, out=0, looseness=1.25] (14) to (7);
		\draw [style=edge_forward, in=90, out=0] (14) to (10);
		\draw [style=edge_forward, in=90, out=0, looseness=1.50] (14) to (11);
		\draw [style=edge_forward, in=180, out=-90, looseness=0.75] (5) to (15);
		\draw [style=edge_forward, in=180, out=-90] (4) to (15);
		\draw [style=edge_forward, in=180, out=270, looseness=1.25] (3) to (15);
		\draw [style=edge_forward, in=180, out=-90, looseness=0.75] (9) to (15);
		\draw [style=edge_forward, in=180, out=-90, looseness=0.50] (8) to (15);
		\draw [style=edge_forward_selected, in=0, out=-90, looseness=1.25] (13) to (15);
		\draw [style=edge_forward, in=0, out=-90, looseness=1.25] (12) to (15);
		\draw [style=edge_forward] (3) to (6);
		\draw [style=edge_forward_selected] (4) to (7);
		\draw [style=edge_forward] (5) to (7);
		\draw [style=edge_forward] (4) to (6);
		\draw [style=edge_forward] (8) to (10);
		\draw [style=edge_forward_selected] (9) to (11);
		\draw [style=edge_forward] (8) to (11);
		\draw [style=edge_forward] (17) to (16);
		\draw [style=edge_forward] (19) to (18);
		\draw [style=edge_forward] (5) to (19);
		\draw [style=edge_forward] (18) to (17);
		\draw [style=edge_forward, in=90, out=0, looseness=0.75] (14) to (19);
		\draw [style=edge_forward, in=180, out=-90] (18) to (15);
		\draw [style=edge_forward, in=90, out=0, looseness=1.75] (14) to (17);
		\draw [style=edge_forward, in=0, out=-90, looseness=1.25] (16) to (15);
		\draw [style=edge_forward] (4) to (19);
		\draw [style=edge_forward] (9) to (17);
	\end{pgfonlayer}
\end{tikzpicture}}
\caption{}
\end{subfigure}

\begin{subfigure}[b]{0.225\textwidth}
\resizebox{0.95\textwidth}{!}{
\begin{tikzpicture}
	\begin{pgfonlayer}{nodelayer}
		\node [style=vertex_default, fill=Red] (0) at (-7, 1) {};
		\node [style=vertex_default] (1) at (-7.5, 0) {};
		\node [style=vertex_default, fill=Red] (2) at (-8, -1) {};
		\node [style=vertex_default, fill=Red] (3) at (-6, 1) {};
		\node [style=vertex_default] (4) at (-6.5, 0) {};
		\node [style=vertex_default, fill=Red] (5) at (-7, -1) {};
		\node [style=vertex_default, fill=Red] (6) at (-4, 1) {};
		\node [style=vertex_default] (7) at (-4.5, 0) {};
		\node [style=vertex_default, fill=Red] (8) at (-3, 1) {};
		\node [style=vertex_default] (9) at (-3.5, 0) {};
		\node [style=vertex_default, fill=Red] (10) at (-1, 1) {};
		\node [style=vertex_default] (11) at (-1.5, 0) {};
		\node [style=vertex_default, fill=Red] (12) at (0, 1) {};
		\node [style=vertex_default] (13) at (-0.5, 0) {};
		\node [style=vertex_default] (14) at (-6, 3) {\LARGE{S}};
		\node [style=vertex_default] (15) at (-2, -3) {\LARGE{T}};
		\node [style=vertex_default, fill=Red] (16) at (-1, -1) {};
		\node [style=vertex_default, fill=Red] (17) at (-2, -1) {};
		\node [style=vertex_default, fill=Red] (18) at (-4, -1) {};
		\node [style=vertex_default, fill=Red] (19) at (-5, -1) {};
	\end{pgfonlayer}
	\begin{pgfonlayer}{edgelayer}
		\draw [style=edge_forward] (0) to (3);
		\draw [style=edge_forward] (2) to (5);
		\draw [style=edge_forward] (6) to (8);
		\draw [style=edge_forward] (10) to (12);
		\draw [style=edge_forward, in=90, out=180, looseness=1.25] (14) to (0);
		\draw [style=edge_forward, in=90, out=180, looseness=1.25] (14) to (2);
		\draw [style=edge_forward, in=90, out=0, looseness=1.25] (14) to (6);
		\draw [style=edge_forward, in=90, out=0, looseness=1.25] (14) to (7);
		\draw [style=edge_forward, in=90, out=0] (14) to (10);
		\draw [style=edge_forward, in=90, out=0, looseness=1.50] (14) to (11);
		\draw [style=edge_forward, in=180, out=-90, looseness=0.75] (5) to (15);
		\draw [style=edge_forward, in=180, out=-90] (4) to (15);
		\draw [style=edge_forward, in=180, out=270, looseness=1.25] (3) to (15);
		\draw [style=edge_forward, in=180, out=-90, looseness=0.75] (9) to (15);
		\draw [style=edge_forward, in=180, out=-90, looseness=0.50] (8) to (15);
		\draw [style=edge_forward, in=0, out=-90, looseness=1.25] (12) to (15);
		\draw [style=edge_forward] (3) to (6);
		\draw [style=edge_forward] (5) to (7);
		\draw [style=edge_forward] (4) to (6);
		\draw [style=edge_forward] (8) to (10);
		\draw [style=edge_forward] (8) to (11);
		\draw [style=edge_forward] (17) to (16);
		\draw [style=edge_forward] (19) to (18);
		\draw [style=edge_forward] (5) to (19);
		\draw [style=edge_forward] (18) to (17);
		\draw [style=edge_forward, in=90, out=0, looseness=0.75] (14) to (19);
		\draw [style=edge_forward, in=180, out=-90] (18) to (15);
		\draw [style=edge_forward, in=90, out=0, looseness=1.75] (14) to (17);
		\draw [style=edge_forward, in=0, out=-90, looseness=1.25] (16) to (15);
		\draw [style=edge_forward] (4) to (19);
		\draw [style=edge_forward] (9) to (17);
		\draw [style=edge_backward, bend right] (14) to (1);
		\draw [style=edge_backward, bend left=45] (1) to (4);
		\draw [style=edge_backward, bend left, looseness=0.75] (4) to (7);
		\draw [style=edge_backward, bend left=45] (7) to (9);
		\draw [style=edge_backward, bend right, looseness=0.75] (9) to (11);
		\draw [style=edge_backward, bend left=45] (11) to (13);
		\draw [style=edge_backward, bend right=330, looseness=0.75] (13) to (15);
	\end{pgfonlayer}
\end{tikzpicture}}
\caption{}
\end{subfigure}
\begin{subfigure}[b]{0.225\textwidth}
\resizebox{0.95\textwidth}{!}{
\begin{tikzpicture}
	\begin{pgfonlayer}{nodelayer}
		\node [style=vertex_default] (0) at (-7, 1) {};
		\node [style=vertex_default] (1) at (-7.5, 0) {};
		\node [style=vertex_default] (2) at (-8, -1) {};
		\node [style=vertex_default] (3) at (-6, 1) {};
		\node [style=vertex_default] (4) at (-6.5, 0) {};
		\node [style=vertex_default] (5) at (-7, -1) {};
		\node [style=vertex_default] (6) at (-4, 1) {};
		\node [style=vertex_default] (7) at (-4.5, 0) {};
		\node [style=vertex_default] (8) at (-3, 1) {};
		\node [style=vertex_default] (9) at (-3.5, 0) {};
		\node [style=vertex_default] (10) at (-1, 1) {};
		\node [style=vertex_default] (11) at (-1.5, 0) {};
		\node [style=vertex_default] (12) at (0, 1) {};
		\node [style=vertex_default] (13) at (-0.5, 0) {};
		\node [style=vertex_default] (14) at (-6, 3) {\LARGE{S}};
		\node [style=vertex_default] (15) at (-2, -3) {\LARGE{T}};
		\node [style=vertex_default] (16) at (-1, -1) {};
		\node [style=vertex_default] (17) at (-2, -1) {};
		\node [style=vertex_default] (18) at (-4, -1) {};
		\node [style=vertex_default] (19) at (-5, -1) {};
	\end{pgfonlayer}
	\begin{pgfonlayer}{edgelayer}
		\draw [style=edge_forward] (0) to (3);
		\draw [style=edge_forward] (2) to (5);
		\draw [style=edge_forward_selected] (6) to (8);
		\draw [style=edge_forward] (10) to (12);
		\draw [style=edge_forward, in=90, out=180, looseness=1.25] (14) to (0);
		\draw [style=edge_forward, in=90, out=180, looseness=1.25] (14) to (2);
		\draw [style=edge_forward_selected, in=90, out=0, looseness=1.25] (14) to (6);
		\draw [style=edge_forward, in=90, out=0, looseness=1.25] (14) to (7);
		\draw [style=edge_forward, in=90, out=0] (14) to (10);
		\draw [style=edge_forward, in=90, out=0, looseness=1.50] (14) to (11);
		\draw [style=edge_forward, in=180, out=-90, looseness=0.75] (5) to (15);
		\draw [style=edge_forward_selected, in=180, out=-90] (4) to (15);
		\draw [style=edge_forward, in=180, out=270, looseness=1.25] (3) to (15);
		\draw [style=edge_forward, in=180, out=-90, looseness=0.75] (9) to (15);
		\draw [style=edge_forward, in=180, out=-90, looseness=0.50] (8) to (15);
		\draw [style=edge_forward, in=0, out=-90, looseness=1.25] (12) to (15);
		\draw [style=edge_forward] (3) to (6);
		\draw [style=edge_forward] (5) to (7);
		\draw [style=edge_forward] (4) to (6);
		\draw [style=edge_forward] (8) to (10);
		\draw [style=edge_forward_selected] (8) to (11);
		\draw [style=edge_forward] (17) to (16);
		\draw [style=edge_forward] (19) to (18);
		\draw [style=edge_forward] (5) to (19);
		\draw [style=edge_forward] (18) to (17);
		\draw [style=edge_forward, in=90, out=0, looseness=0.75] (14) to (19);
		\draw [style=edge_forward, in=180, out=-90] (18) to (15);
		\draw [style=edge_forward, in=90, out=0, looseness=1.75] (14) to (17);
		\draw [style=edge_forward, in=0, out=-90, looseness=1.25] (16) to (15);
		\draw [style=edge_forward] (4) to (19);
		\draw [style=edge_forward] (9) to (17);
		\draw [style=edge_backward, bend right] (14) to (1);
		\draw [style=edge_backward, bend left=45] (1) to (4);
		\draw [style=edge_backward_selected, bend left, looseness=0.75] (4) to (7);
		\draw [style=edge_backward_selected, bend left=45] (7) to (9);
		\draw [style=edge_backward_selected, bend right, looseness=0.75] (9) to (11);
		\draw [style=edge_backward, bend left=45] (11) to (13);
		\draw [style=edge_backward, bend right=330, looseness=0.75] (13) to (15);
	\end{pgfonlayer}
\end{tikzpicture}}
\caption{}
\end{subfigure}
\end{center}

\begin{center}
\begin{subfigure}[b]{0.225\textwidth}
\resizebox{0.95\textwidth}{!}{
\begin{tikzpicture}
	\begin{pgfonlayer}{nodelayer}
		\node [style=vertex_default] (0) at (-7, 1) {};
		\node [style=vertex_default] (1) at (-7.5, 0) {};
		\node [style=vertex_default] (2) at (-8, -1) {};
		\node [style=vertex_default] (3) at (-6, 1) {};
		\node [style=vertex_default] (4) at (-6.5, 0) {};
		\node [style=vertex_default] (5) at (-7, -1) {};
		\node [style=vertex_default, fill=Blue] (6) at (-4, 1) {};
		\node [style=vertex_default, fill=Red] (7) at (-4.5, 0) {};
		\node [style=vertex_default, fill=Blue] (8) at (-3, 1) {};
		\node [style=vertex_default, fill=Red] (9) at (-3.5, 0) {};
		\node [style=vertex_default] (10) at (-1, 1) {};
		\node [style=vertex_default] (11) at (-1.5, 0) {};
		\node [style=vertex_default] (12) at (0, 1) {};
		\node [style=vertex_default] (13) at (-0.5, 0) {};
		\node [style=vertex_default] (14) at (-6, 3) {\LARGE{S}};
		\node [style=vertex_default] (15) at (-2, -3) {\LARGE{T}};
		\node [style=vertex_default] (16) at (-1, -1) {};
		\node [style=vertex_default] (17) at (-2, -1) {};
		\node [style=vertex_default, fill=Red] (18) at (-4, -1) {};
		\node [style=vertex_default, fill=Blue] (19) at (-5, -1) {};
	\end{pgfonlayer}
	\begin{pgfonlayer}{edgelayer}
		\draw [style=edge_forward] (0) to (3);
		\draw [style=edge_forward] (2) to (5);
		\draw [style=edge_backward, bend left=45] (6) to (8);
		\draw [style=edge_forward] (10) to (12);
		\draw [style=edge_forward, in=90, out=180, looseness=1.25] (14) to (0);
		\draw [style=edge_forward, in=90, out=180, looseness=1.25] (14) to (2);
		\draw [style=edge_backward, in=165, out=270] (14) to (6);
		\draw [style=edge_forward, in=90, out=0, looseness=1.25] (14) to (7);
		\draw [style=edge_forward, in=90, out=0] (14) to (10);
		\draw [style=edge_forward, in=90, out=0, looseness=1.50] (14) to (11);
		\draw [style=edge_forward, in=180, out=-90, looseness=0.75] (5) to (15);
		\draw [style=edge_backward, in=180, out=-60] (4) to (15);
		\draw [style=edge_forward, in=180, out=270, looseness=1.25] (3) to (15);
		\draw [style=edge_forward, in=180, out=-90, looseness=0.75] (9) to (15);
		\draw [style=edge_forward, in=180, out=-90, looseness=0.50] (8) to (15);
		\draw [style=edge_forward, in=0, out=-90, looseness=1.25] (12) to (15);
		\draw [style=edge_forward] (3) to (6);
		\draw [style=edge_forward] (5) to (7);
		\draw [style=edge_forward] (4) to (6);
		\draw [style=edge_forward] (8) to (10);
		\draw [style=edge_backward, in=191, out=-79, looseness=0.75] (8) to (11);
		\draw [style=edge_forward] (17) to (16);
		\draw [style=edge_forward] (19) to (18);
		\draw [style=edge_forward] (5) to (19);
		\draw [style=edge_forward] (18) to (17);
		\draw [style=edge_forward, in=90, out=0, looseness=0.75] (14) to (19);
		\draw [style=edge_forward, in=180, out=-90] (18) to (15);
		\draw [style=edge_forward, in=90, out=0, looseness=1.75] (14) to (17);
		\draw [style=edge_forward, in=0, out=-90, looseness=1.25] (16) to (15);
		\draw [style=edge_forward] (4) to (19);
		\draw [style=edge_forward] (9) to (17);
		\draw [style=edge_backward, bend right] (14) to (1);
		\draw [style=edge_backward, bend left=45] (1) to (4);
		\draw [style=edge_forward] (4) to (7);
		\draw [style=edge_forward] (7) to (9);
		\draw [style=edge_forward] (9) to (11);
		\draw [style=edge_backward, bend left=45] (11) to (13);
		\draw [style=edge_backward, bend right=330, looseness=0.75] (13) to (15);
	\end{pgfonlayer}
\end{tikzpicture}}
\caption{}
\end{subfigure}
\end{center}
\caption{An illustration for 2-pass DP with quadratic interactions. (a) the
initial DAG graph, a pair of nodes indicate a candidate detection; (b) first
iteration of the algorithm, red edges indicates the shortest path found in this
iteration; (c) we reverse all the edges along the shortest path (green edges), and
add the pairwise cost imposed by this path to other candidates detections at
each time point (red node pairs). (d) In the second iteration of the algorithm, red 
edges and blue edges indicate the new shortest path which happens to traverse
three of the reversed edges (shown in blue); (e) we again reverse all the edges 
along the shortest path yielding two instanced st-tracks (green edges). We
again update pairwise cost: blue node pair indicates we subtract the pairwise cost 
imposed by ``turning off'' a candidate, red pair indicates adding in pairwise cost 
of newly instanced candidates, and the blue-red pair indicates we first add the pairwise
cost by newly instanced candidates, then subtract the pairwise cost by newly
uninstanced candidates. Additions and subtractions are done to the non-negated
edge costs and then negated if necessary.}
\label{fig:2passupdate}
\end{figure}
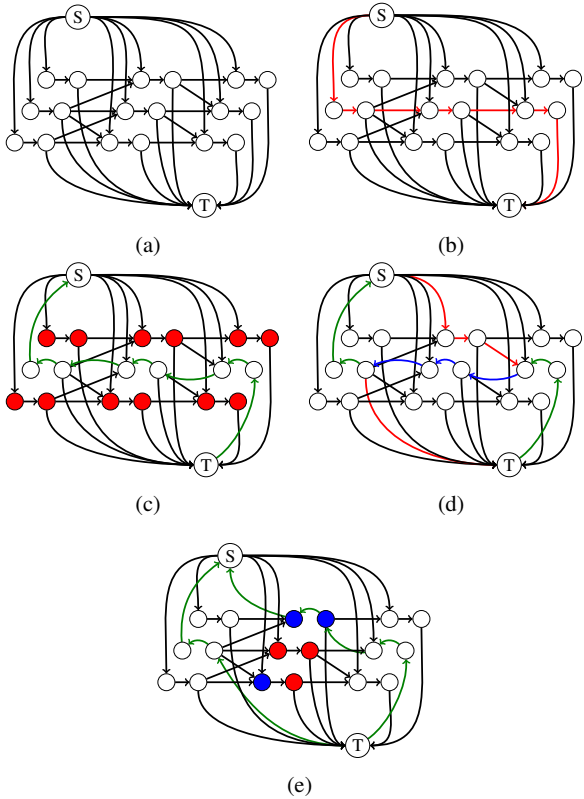

\paragraph{Quadratic cost updates}
The path found by 2-pass DP may contain both forward and backward detection edges
which correspond to newly instanced and removed detections respectively. When we 
augment the flow with this new path, we also
update the (unary) cost of other nodes by adding or subtracting the pairwise cost imposed by
turning on or off selected nodes on the path.
The entire procedure is described as Algorithm~\ref{alg:alg2} and illustrated in 
Figure \ref{fig:2passupdate}.
\begin{algorithm}{}
\begin{minipage}{0.45\textwidth}
\caption{Two-pass DP with Quadratic Cost Update}
\label{alg:alg2}
\begin{algorithmic}[1]
\State \textbf{Input}: A Directed-Acyclic-Graph $G$ with node and edge weights $c_i, c_{ij}$
\State initialize $\mathbf{f} = \mathbf{0}$
\Repeat
  \State Find a shortest st-path $\mathbf{p}$ in $G_r(\mathbf{f})$ using 2-pass DP
  \State $track\_cost = cost(\mathbf{p})$
  \If {$track\_cost < 0$}
    \ForAll{locations $x_i \in \mathbf{p}$}
      \If {$f_i = 0$}
        \State $c_j = c_j + q_{ij} + q_{ji}, \forall ij \in EC$
      \Else
        \State $c_j = c_j - q_{ij} - q_{ji}, \forall ij \in EC$
      \EndIf
    \EndFor
    \State $\mathbf{f}(\mathbf{p}) = \mathbf{1} - \mathbf{f}(\mathbf{p})$
  \EndIf
\Until{$track\_cost \ge 0$}
\State \textbf{Output}: Solution $\mathbf{f}$
\end{algorithmic}
\end{minipage}
\end{algorithm}

Notice that to simplify our notation, we construct temporary residual graph at
the beginning of each iteration based on the current costs. In practice, we 
instead update edge costs and directions on the original graph at the end of each 
iteration. When operating in place on the residual graph, the signs of the cost
updates are reversed when updating costs of reversed detection edges
(i.e., when turning off a detection, we subtract pairwise costs from forward detection
edges but add pairwise costs to reversed detection edges).

\subsubsection{Caching DP messages}
\label{sec:caching}
Similar to the speed-up techniques employed by~\cite{PirsiavashRF_CVPR_2011},
DP algorithms with contextual updates only need to re-evaluate a subset of
all detection nodes in each round of shortest path computation.  For one-pass DP, we need to re-evaluate detection nodes who have the same birth node as either newly found shortest path or suppressed
nodes, \ie, nodes whose cost has been increased due to most recent contextual update.
Then, we also need to count the effect of boosted nodes whose
cost has been decreased due to most recent contextual update; this is done by first setting
all boosted nodes to be active, re-evaluating their successors and propagating this
activity to nodes whose $link$ points back to the active nodes. The caching scheme is 
similar for two-pass DP, only difference being that we have to conduct two forward passes 
and one backward pass, thus we need to maintain separate caches for each pass.

\subsubsection{Time Complexity Analysis}
\label{sec:complexity}
For the basic network-flow problem in Eq.~\ref{eqn:mincostflow} with $n$ total
variables in $\mathbf{f}$ and $k$ detections, exact successive shortest path
using Dijkstra's algorithm has a worst-case performance of $O(n \log k)$
operations per path and terminates after adding fewer than $k$ paths yielding
$O(n k \log k)$ worst-cast performance. The family of DP algorithms introduced
by \cite{PirsiavashRF_CVPR_2011} takes $O(n)$ to find a single track and thus
has worst-case performance of $O(kn)$ for basic network-flow problem.

For solving the linear program in Eq. ~\ref{eqn:lprelax}, a general solver such
as simplex has average run times of $O(n^3)$, where $n$ is the total number of
variables (unary and pairwise). For a video with $k$ total detections, our
one-pass DP algorithm takes $O(n)$ to find a single track, achieving a worst
case $O(kn)$ time complexity. Notice that $k$ is often much smaller than $n$;
in fact, for short sequences $n$ often grows quadratically with $k$. The same
$O(kn)$ worst-case time complexity applies for 2-pass DP or any number of
constant passes because the complexity of every iteration of multi-pass DP still
scales linearly with $n$.

\section{Features for Scoring Tracks}
\label{sec:features}

In order to learn the tracking potentials ($\mathbf{c}$ and $\mathbf{q}$) we
parameterize the flow cost objective by a vector of weights $\mathbf{w}$ and a set of
features $\Psi(X,\mathbf{f})$ that depend on features extracted from the video,
the spatio-temporal relations between candidate detections, and which tracks
are instanced.  With this linear parameterization we write the cost of a given
flow as $C(\mathbf{f}) = \mathbf{w}^T \Psi(X,\mathbf{f})$ where the vector
components of the weight and feature vector are given by:
\begin{align}
\mathbf{w} =
\begin{bmatrix}
w_S\\ w_t\\ w_s\\ w_a\\ w_E
\end{bmatrix}
\indent \Psi(X,\mathbf{f}) = 
\begin{bmatrix}
\sum_i \phi_S(x_i^s) f_i^s \\ \sum_{ij \in E} \psi_t(x_i, x_j) f_{ij} \\ \sum_{ij \in EC} \psi_s(x_i, x_j) f_i f_j \\ \sum_i \phi_a(x_i) f_i \\ \sum_i \phi_E(x_i^t) f_i^t
\end{bmatrix}
\end{align}
Here $w_a$ represents local appearance template for the tracked objects of
interest, $w_t$ represents weights for transition features, $w_s$ represents
weights for pairwise interactions, $w_S$ and $w_E$ represents weights associated
with track births and deaths. $\Psi(X,\mathbf{f})$ are corresponding features.
Given the wegiht vector $\mathbf{w}$, we extract features on each node $i$,
including detections $\phi_a(x_i)$ and track birth/deaths
$\phi_S(x_i^s)$,$\phi_E(x_i^t)$, along with features on each edge $ij \in E$ and
$ij \in EC$, including transitions $\psi_t(x_i, x_j)$ and pairwise interactions
$\psi_s(x_i, x_j)$. Then we multiply corresponding weight vectors and features
on each edge. In this way we can obtain costs of each node/edge in the network
and conduct standard inference as described in Section~\ref{sec:inference}. We
describe each type of features as below:

\paragraph{Local appearance and birth/death model} We make use of
off-the-shelf detectors~(\cite{DollarPAMI14pyramids,voc-release4,Wang_2013_ICCV})
to capture local appearance.  Our local appearance
feature thus consists of the detector score along with a constant 1 to allow
for a variable bias.  In applications with static cameras it can be useful to
learn a spatially varying bias to model where tracks are likely to appear or
disappear.  However, most videos in our experiments are captured from moving
platforms, we thus use a single constant value 1 for the birth and death
features.

\paragraph{Transition model} We connect a candidate $x_i$ at time $t_i$ with
another candidate $x_j$ at a later time $t_i+n$, only if the overlap ratio
between $x_i$'s window and $x_j$'s window exceeds $0.3$. The overlap ratio is
defined as two windows' intersection over their union. We use this overlap ratio
to compute a binary overlap feature associated with each transition link which
is 1 if the overlap ratio is lower than 0.5, and 0 otherwise.  In order to
handle occlusion, we allow up to 8 frames(10 on PETS+TUD-Stadtmitte) gap between
the two detection sites of a transition edge. We jointly encode the overlap and
frame gap with a single 16 dimensional (20 on PETS+TUD-Stadtmitte) binary
feature for each transition link.

\paragraph{Pairwise interactions}
The weight vector $w_s$ encodes valid geometric configurations of two
 tracked objects in a frame. $\psi_s(x_i, x_j)$ is a discretized spatial-context feature
 that bins relative location of detection window at location $x_i$ and window at
 location $x_j$ into one of the $D$ relations including on top of, above, below,
 next-to, near, far and overlap (similar to the spatial context
 by~\cite{DesaiRF_ICCV_2009}). To mimic the temporal NMS described by 
 \cite{PirsiavashRF_CVPR_2011} we add one additional relation, strict overlap,
 which is set to 1 if the ratio of the intersection area of two boxes over
 the area of the first box is greater than 0.9.
 If we assume that there are $K$ classes of objects in
 the video, then $w_s$ is a $DK^2$ vector, $w_s = [w_{s11}^T, w_{s12}^T,
 ..., w_{sij}^T, ... , w_{sKK}^T]^T$, in which $w_{sij}$ is a length of $D$
 column vector that encodes valid geometric configurations of object of class
 $i$ w.r.t. object of class $j$. This allows the model to capture intra- and
 inter-class contextual relationships between tracks.

\section{Learning}
\label{sec:learning}

We formulate parameter learning of tracking models as a structured prediction
problem.  With some abuse of notation, assume we have $N$ labeled training videos
$(X_n, \mathbf{f}_n) \in \mathcal{X} \times \mathcal{F}$ indexed by $n = 1,..., N$. Given
ground-truth tracks in training videos specified by flow variables $\mathbf{f}_n$,
we discriminatively learn tracking model parameters $\mathbf{w}$ using a
structured SVM with margin scaling:
\begin{align}
&\mathbf{w}^* = \underset{\mathbf{w},\xi_n \ge 0}{\operatorname{argmin}} \ \frac{1}{2} \| \mathbf{w} \| ^2  + C \sum_n \xi_n \label{eqn:structsvm}\\
\text{s.t.} &\ 
\mathbf{w}^T \Psi(X_n,\widehat{\mathbf{f}}) - \mathbf{w}^T \Psi(X_n,\mathbf{f}_n) \ge L(\mathbf{f}_n, \widehat{\mathbf{f}}) - \xi_n \nonumber
\quad \forall n, \widehat{\mathbf{f}}
\end{align}
$\Psi(X_n,\mathbf{f}_n)$ are the features extracted from $n$th training video.
$L(\mathbf{f}_n, \widehat{\mathbf{f}})$ is a loss
function that penalizes any difference between the inferred label
$\widehat{\mathbf{f}}$ and the ground truth label $\mathbf{f}_n$ and
which satisfies $L(\mathbf{f},\mathbf{f})=0$. The
constraint on the slack variables $\xi_n$ ensure that we pay a penalty for
any training videos in which the cost of the flow associated with ground-truth
tracks under model $\mathbf{w}$ is higher than some other incorrect flow
$\widehat{\mathbf{f}}$.

\subsection{Cutting Plane Optimization}
We optimize the structured SVM objective in Eq.~\ref{eqn:structsvm} using a
standard cutting-plane method~(\cite{Joachims/etal/09a}) in which the exponential
number of constraints (one for each possible flow $\widehat{\mathbf{f}}$)
are approximated by a much smaller number of terms.  Given a current 
estimate of $\mathbf{w}$ we find a ``most violated constraint'' for each training
video:
\begin{align}
\widehat{\mathbf{f}}_n^* &= \underset{\widehat{\mathbf{f}}}{\operatorname{argmin}} \ \langle \mathbf{w},
\Psi(X_n,\widehat{\mathbf{f}}) - \Psi(X_n,\mathbf{f_n}) \rangle - L(\mathbf{f_n}, \widehat{\mathbf{f}}) 
\label{eqn:loss_augmented}
\end{align}
We then add constraints for the flows $\{\widehat{\mathbf{f}}_n^*\}$ to the optimization problem
and solve for an updated $\mathbf{w}$.  This procedure is iterated until no
additional constraints are added to the problem.  In our implementation, at
each iteration we add a single linear constraint which is a sum of violating
constraints derived from individual videos in the dataset. This linear 
combination is also a valid cutting plane constraint~(\cite{DesaiRF_ICCV_2009})
and yields faster overall convergence.

The key subroutine is finding the most-violated constraint for a given video which
requires solving the loss-augmented inference problem Eq.~\ref{eqn:loss_augmented}.
As long as the loss function $L(\mathbf{f}, \widehat{\mathbf{f}})$ 
decomposes as a sum over flow variables then this problem has the same form
as our test-time tracking inference problem, the only difference being that the
cost of variables in $\mathbf{f}$ is augmented by their corresponding negative
loss.

We note that this formulation allows for constraints corresponding
to non-integral flows $\widehat{\mathbf{f}}$ so we can directly use the LP
relaxation (Eq. \ref{eqn:lprelax}) to generate violated constraints during
training.  \cite{Finley/Joachims/08a} point out that besides optimality
guarantees, including non-integral constraints naturally pushes the SVM
optimization towards model parameters that produce integer solutions even
before rounding.

\subsection{Tracking Loss Function}
We find that a critical aspect for successful learning is to use a loss
function that closely resembles major tracking performance criteria,
such as Multiple Object Tracking Accuracy (MOTA). Metrics such as false
positive, false negative, true positive, true negative and true/false birth/death
can be easily incorporated using a standard Hamming loss on the flow vector.
However, id switches and fragmentations are determined by
looking at labels of two consecutive transition links simultaneously and hence
cannot be optimized by our inference routine (which only considers pairwise
relations between detections within a frame).  Instead, we propose a
decomposable loss for transition links that attempts to capture important
aspects of MOTA by taking into account the length and localization of
transition links rather than simply using a constant (Hamming) loss on
mislabled links.

We define a weighted Hamming loss to measure distance between ground-truth
tracks $\mathbf{f}$ and inferred tracks $\widehat{\mathbf{f}}$ that includes
detections/birth/death, $f_i$, and transitions, $f_{ij}$.  Let
\[
L(\widehat{\mathbf{f}} , \mathbf{f}) = 
\sum_i l_i \left| f_i -  \widehat{f_i} \right| + \sum_{ij} l_{ij} \left| f_{ij} -  \widehat{f_{ij}} \right|
\]
where $\mathbf{l} = \{ l_1, ..., l_i,...,l_{ij},...,l_{|\mathbf{f}|} \}$ is a
vector indicating the penalty for differences between the estimated
flow $\widehat{\mathbf{f}}$ and the ground-truth $\mathbf{f}$.

\begin{figure}[t]
\begin{center}
\includegraphics[clip,trim=0cm 0cm 0cm 0cm,width=0.47\textwidth,height=0.27\textwidth]{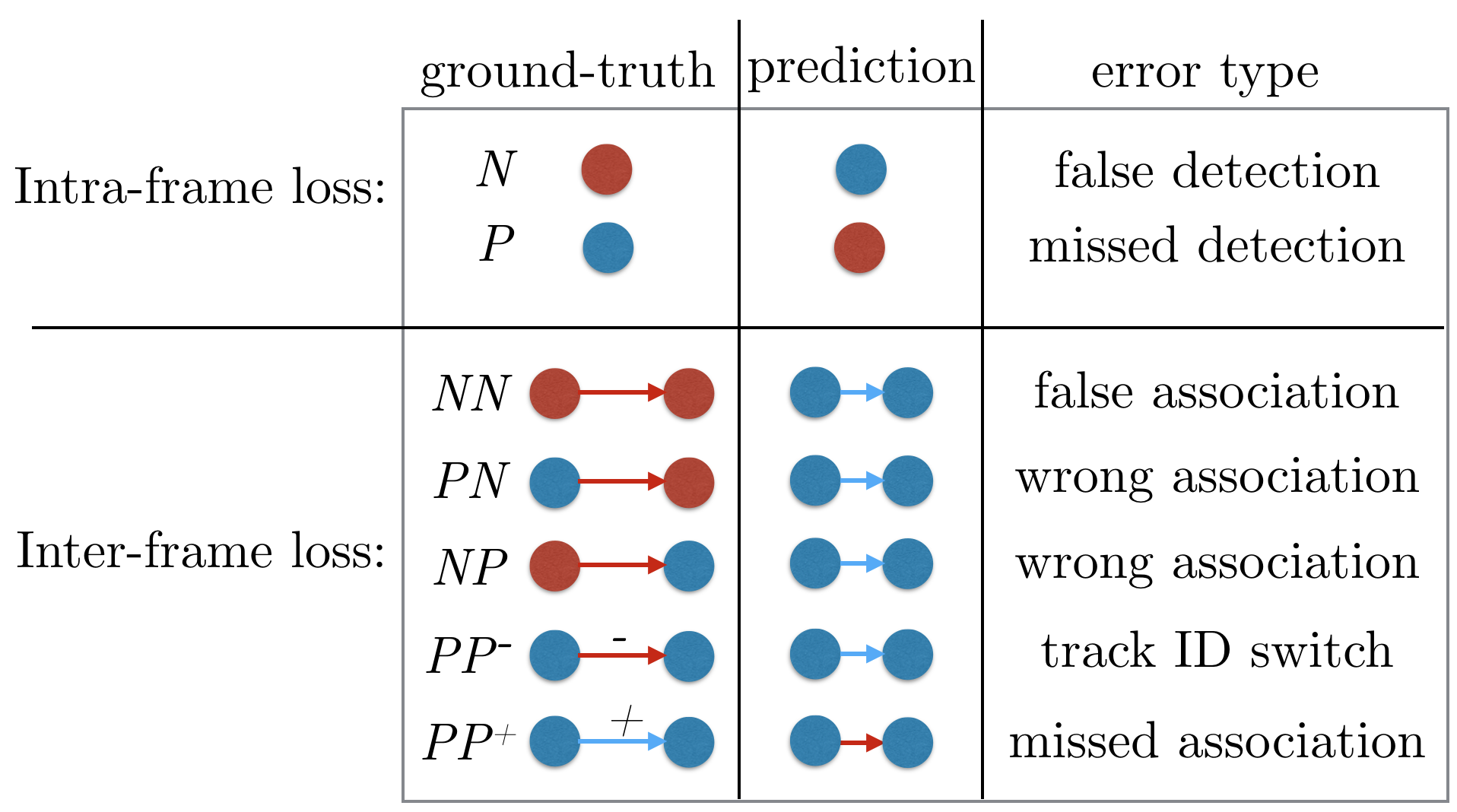}
\end{center}
\caption{Possible intra-frame and inter-frame errors a tracker could make which are 
penalized by our loss function. A blue node/edge indicates that corresponding flow variable
is set to 1 in either ground-truth or tracker prediction. Similarly red node/edge indicates 
corresponding flow variable is set to 0.}
\label{fig:loss}
\end{figure}

In order to describe our transition loss, let us first denote four types of
transition links present in the tracking graph: $NN$ is the link from a false
detection to another false detection, $PN$ is the link from a true detection to
a false detection, $NP$ is the link from a false detection to a true detection,
$PP^+$ is the link from a true detection to another true detection with the
same identity, and $PP^-$ is the link from a true detection to another true
detection with a different identity. These are depicted visually in Figure
\ref{fig:loss}.

For all the transition links with a frame gap larger than 1, we interpolate
detections between its start detection and end detection.  The interpolated
``virtual detections'' are considered to be either true virtual detection or
false virtual detection, depending on whether they overlap with a ground truth
label or not.  We define the losses for different types of transitions as:

\begin{enumerate}
  \item $NN$ :  $\; l_{ij} =$ ( \# true virt. det. + \# false virt. det.)
  \item $PN$ :  $\; l_{ij} =$ (\# true virt. det. + \# false virt. det. + 1)
  \item $NP$ :  $\; l_{ij} =$ (\# true virt. det. + \# false virt. det. + 1)
  \item $PP^-$ : $\; l_{ij} =$ (\# true virt. det. + \# false virt. det. + 2)
  \item $PP^+$ : $\; l_{ij} =$ (\# true virt. det.)
\end{enumerate}

\subsection{Ground-truth flows from training data}
Available training datasets specify ground-truth bounding boxes that need to be
mapped onto ground-truth flow variables $\mathbf{f_n}$ for each video.  To do
this mapping, we first consider each frame separately. We take the highest
scoring detection window that overlaps a ground truth label as true detection
and assign it a track identity label which is the same as the ground truth
label it overlaps.  Next, for each track identity, we run a simplified version
of the dynamic programming algorithm to find the path that claims the largest
number of true detections.  After we iterate through all id labels, any
instanced graph edge will be a true detection/transition/birth/death while the
remainder will be false.

An additional difficulty of training which arises on the KITTI tracking
benchmark is special evaluation rules for ground truth labels such as
small/truncated objects and vans for cars, sitting persons for pedestrians.
This is resolved in our training procedure by removing all detection candidates
that correspond to any of these ``ambiguous'' ground truth labels during
training; in this way we avoid mining hard negatives from those labels. 
To speed up training on both MOT and KITTI dataset, we partition full-sized
training sequences into 10-frame-long subsequences with a 5-frame overlap, and
define losses on each subsequence separately.

\section{Experimental results}
\label{sec:experiments}

Historically it has been challenging to make a meaningful empirical comparison
among tracking performance results reported in the literature as the exact
detection set, evaluation script, amount of training data and even ground-truth
labels have varied greatly. As a result, there has been significant recent
efforts to establish standard benchmarks for evaluating multi-target tracking
algorithms. We focus our diagnostic analysis on the KITTI and MOT Challenge
datasets as they have very clear train-test splits of data and the ground-truth
labels for testing data are strictly ``held out'', invisible to all competitors
of the benchmark. To make our results easy to compare to other algorithms, we
always use the base detections, ground-truth and evaluation scripts provided by
the MOT and KITTI benchmark organizers when performing training, inference and
diagnostics.

To aid comparison to older methods that have not been evaluted on MOT or KITTI,
we also include results on PETS and the TUD-Stadtmitte sequence, based on what
we believe to be the most popular public detection set, ground-truth and
evaluation script.  However, we note that for these videos the evaluation is
less standardized and since ground-truth annotations have been used widely to
tune models, there may be overfitting. We thus view these results as less
informative than those on MOT and KITTI test benchmarks.

\subsection{Datasets and Benchmarks}
The Multiple Object Tracking
Benchmark\footnote{\url{http://nyx.ethz.ch/}}~(\cite{MOTChallenge2015}) targets
primarily pedestrian tracking. It contains sequences from widely-used
multi-target tracking benchmarks such as TUD, ETH, PETS09, TownCentre and
KITTI, augmented with several additional newly acquired sequences. The dataset
is split into 11 training sequences and 11 testing sequences, with ground-truth
labels of testing set held out on a private server.

The KITTI Tracking
Benchmark\footnote{\url{http://www.cvlibs.net/datasets/kitti/eval_tracking.php}}~(\cite{Geiger2012CVPR})
involves multi-category tracking of cars, pedestrians and cyclists. It consists
of 21 training sequences and 29 testing sequences which is much larger than the
MOT dataset.  Similar to the MOT benchmark, the ground-truth labels for testing
set are also held out.  Labels for the cyclist category are available on the
training set but the benchmark server does not provide test-set benchmarks for
cyclist. For both MOT and KITTI we allow up to 8 frames gap between two
detection sites of a transition edge and determined the best regularization
parameters via leave-one-video-out cross-validation on the training data.

Finally, in order to compare to previous related work such
as~\cite{BrauICCV2013} and \cite{ChariCVPR15}, we use data provided
by~\cite{Milan:2013:DTE}\footnote{\url{http://www.milanton.de/data/}}
consisting of 5 sequences from PETS09 dataset and the single TUD-Stadtmitte
sequence.  For this data, we report leave-one-video-out cross-validation
results. We allow up to 10 frames gap between two detection sites of a transition
edge in this setup and partition videos into 15-frame long sequences during
training stage. 

\subsection{Evaluation Metrics}
To evaluate the performance of each proposed tracker, we employ a standard
battery of performance measure, which consists of the popular CLEAR MOT metric
(\cite{Bernardin:2008:EMO:1384968.1453688}) and the Mostly-Tracked/Mostly-Lost
metric~(\cite{Li09learningto}):

\begin{itemize}
  \item $MOTA(\uparrow)$: Multi-object Tracking Accuracy.
  \item $MOTP(\uparrow)$: Multi-object Tracking Precision.
  \item $MT(\uparrow)$: ratio of mostly-tracked ground-truth trajectories.
  \item $ML(\downarrow)$: ratio of mostly-lost ground-truth trajectories.
  \item $IDSW(\downarrow)$: total number of identity-switches.
  \item $FRAG(\downarrow)$: total number of times ground-truth trajectories are interrupted.
\end{itemize}

For measurements with $(\uparrow)$, higher scores indicate better performance;
for measurements with $(\downarrow)$, lower scores indicate better performance.

\begin{table*}[tbp]
\begin{center}
\renewcommand{\tabcolsep}{3.0pt}
{\small
\begin{tabular}{ c | c | c | c | c | c | c | c | c | c | c | c | c | c | }
\multicolumn{14}{c}{MOT dataset} \\
\end{tabular}\\
\begin{tabular}[t]{ c | c | c | c | c | c | c | }
\multicolumn{7}{c |}{Benchmark on MOT test set}\\
\hline
Method & MOTA & MOTP & MT & ML & IDSW & FRAG \\
\hline
\rowcolor{Gray}
\cite{Bae_2014_CVPR} & 15.1\% & 70.5\% & 3.2\% & 55.8\% & 637 & 1716 \\
\rowcolor{Gray}
\cite{DBLP:conf/wacv/YoonYLY15} & 18.6\% & 69.6\% & 5.3\% & 53.3\% & 684 & 1282 \\
\cite{Milan:2016:PAMI} & 19.6\% & 71.4\% & 5.1\% & 54.9\% & 521 & 819 \\
\cite{Milan:2015:CVPR}* & 22.5\% & 71.7\% & 5.8\% & 63.9\% & 697 & 737 \\
\cite{LeaFen2014}* & 23.1\% & 70.9\% & 4.7\% & 52.0\% & 1018 & 1061 \\
\rowcolor{Gray}
\cite{XiangICCV15} & 30.3\% & 71.3\% & 13.0\% & 38.4\% & 680 & 1500\\
\rowcolor{Gray}
\cite{KimICCV15} & 32.4\% & 71.8\% & 16.0\% & 43.8\% & 435 & 826\\
\rowcolor{Gray}
\cite{ChoiICCV15} & 33.7\% & 71.9\% & 12.2\% & 44.0\% & 442 & 823\\
\hline
Ours(LP) & 25.2\% & 71.7\% & 5.8\% & 53.0\% & 646 & 849 \\
\hline
\end{tabular}
\begin{tabular}[t]{ | c | c | c | c | c | c | c | }
\multicolumn{7}{c |}{Cross-validation on MOT training set} \\
\hline
Method & MOTA & MOTP & MT & ML & IDSW & FRAG \\
\hline
SSP & \bb{28.7}\% & 72.9\% & 15.1\% & 50.5\% & \bb{440} & \bb{541} \\
\hline
LP+Hamming & 25.3\% & 72.4\% & \bb{17.4}\% & \bb{46.5}\% & 567 & 604 \\
LP & 28.5\% & 72.8\% & 15.1\% & 48.9\% & 440 & 563 \\
DP1 & 27.6\% & 72.4\% & 15.5\% & 49.1\% & 492 & 626 \\
DP2 & 28.5\% & \bb{72.9}\% & 15.5\% & 48.9\% & 476 & 592 \\
\hline
\end{tabular}
}
\end{center}
\vspace{0.05in}
\caption{Benchmark and cross-validation results on MOT dataset. Gray background
indicates the method uses target-specific appearance model, while asterisk(*) indicates
method uses image-evidence, method with neither gray background nor asterisk
can operate purely on bounding boxes without accessing images. We denote
variants of our model as follows: 1) SSP is a model without any pairwise cost terms,
learned and tested with successive shortest path algorithm. 2) LP are models
with pairwise terms, learned with LP-Relaxation while tested with LP-Rounding.
3) DP1 and DP2 are models learned with LP-Relaxation and tested
with 1-pass and 2-pass DP respectively. 4) LP+Hamming is the same as LP, except that
models are learned using Hamming loss instead of the loss described in Section
\ref{sec:learning}.
}
\label{tab:tab1}
\end{table*}
\begin{table*}[tbp]
\begin{center}
\renewcommand{\tabcolsep}{3.0pt}
{\small
\begin{tabular}{ c | c | c | c | c | c | c | c | c | c | c | c | c | c | c | c | c | c |}
\multicolumn{14}{c}{Benchmark on KITTI test set (as of 05/24/2016)} \\
\hline
\multicolumn{7}{c |}{Benchmark on Car, DPM detections} & \multicolumn{7}{c |}{Benchmark on Pedestrian, DPM detections} \\
\hline
Method & MOTA & MOTP & MT & ML & IDSW & FRAG & Method & MOTA & MOTP & MT & ML & IDSW & FRAG \\
\hline
\rowcolor{Gray}
\cite{Geiger2014PAMI} & 54.2\% & 78.4\% & 13.9\% & 34.3\% & 31 & 535 & \cite{Geiger2014PAMI} & NA & NA & NA & NA & NA & NA \\
\cite{Milan:2014:CEM} & 50.2\% & 77.1\% & 14.5\% & 34.0\% & 125 & 398 & \cite{Milan:2014:CEM} & 27.4\% & 68.5\% & 7.9\% & 52.9\% & 96 & 610 \\
\rowcolor{Gray}
\cite{DBLP:conf/wacv/YoonYLY15} & 51.5\% & 75.2\% & 15.2\% & 33.5\% & 51 & 382 & \cite{DBLP:conf/wacv/YoonYLY15} & 34.5\% & 68.1\% & 10.0\% & 47.4\% & 81 & 692 \\
\rowcolor{Gray}
\cite{ChoiICCV15} & 65.2\% & 78.2\% & 31.6\% & 27.9\% & 13 & 154 & \cite{ChoiICCV15} & 36.9\% & 67.8\% & 14.4\% & 42.6\% & 34 & 800 \\
\hline
Ours(LP) & 60.5\% & 76.9\% & 27.7\% & 23.8\% & 16 & 430 & Ours(LP) & 33.3\% & 67.4\% & 9.6\% & 45.0\% & 72 & 825 \\
\hline \hline
\multicolumn{7}{c |}{Benchmark on Car, Regionlet detections} & \multicolumn{7}{c |}{Benchmark on Pedestrian, Regionlet detections} \\
\hline
\rowcolor{Gray}
\cite{DBLP:conf/wacv/YoonYLY15} & 65.3\% & 75.4\% & 26.8\% & 11.4\% & 215 & 742 & \cite{DBLP:conf/wacv/YoonYLY15} & 43.7\% & 71.0\% & 16.8\% & 41.2\% & 156 & 760 \\
\rowcolor{Gray}
\cite{ChoiICCV15} & 77.8\% & 79.5\% & 43.1\% & 14.6\% & 36 & 225 & \cite{ChoiICCV15} & 46.4\% & 71.5\% & 23.4\% & 34.7\% & 63 & 672 \\
\hline
Ours(LP) & 77.2\% & 77.8\% & 43.1\% & 9.0\% & 63 & 558 & Ours(LP) & 43.8\% & 70.5\% & 16.8\% & 34.7\% & 73 & 814 \\
\hline
\end{tabular}}
\end{center}
\begin{center}
\renewcommand{\tabcolsep}{3.0pt}
{\small
\begin{tabular}{ c | c | c | c | c | c | c || c | c | c | c | c | c | c | c | c | c | c |}
\multicolumn{14}{c}{Cross-validation result on KITTI training set} \\
\hline
\hline
\multicolumn{7}{c ||}{Benchmark on Car, DPM detections} & \multicolumn{7}{c |}{Benchmark on Car, Regionlet detections} \\
\hline
Method & MOTA & MOTP & MT & ML & IDSW & FRAG & Method & MOTA & MOTP & MT & ML & IDSW & FRAG \\
\hline
SSP & 64.9\% & \bb{77.9}\% & 27.3\% & 19.6\% & \bb{3} & \bb{186} & SSP & 80.5\% & 80.1\% & 44.4\% & 7.9\% & \bb{17} & \bb{293} \\
\hline
LP & 65.4\% & 77.6\% & 29.6\% & 18.3\% & 4 & 215 & LP & \bb{81.0}\% & \bb{80.1}\% & 44.3\% & 7.2\% & 23& 305 \\
DP1 & \bb{66.0}\% & 77.4\% & 30.5\% & 18.3\% & 15 & 218 & DP1 & 79.0\% & 79.5\% & 44.1\% & 7.1\% & 149 & 509 \\
DP2 & 65.7\% & 77.6\% & \bb{30.7}\% & \bb{18.3}\% & 4 & 203 & DP2 & 80.7\% & 80.0\% & \bb{44.4}\% & \bb{7.2}\% & 62 & 360 \\
\hline \hline
\multicolumn{7}{c ||}{Benchmark on Pedestrian, DPM detections} & \multicolumn{7}{c |}{Benchmark on Pedestrian, Regionlet detections} \\
\hline
Method & MOTA & MOTP & MT & ML & IDSW & FRAG & Method & MOTA & MOTP & MT & ML & IDSW & FRAG \\
\hline
SSP & 49.7\% & \bb{72.8}\% & 19.2\% & 24.0\% & \bb{22} & \bb{231} & SSP & 71.8\% & 76.1\% & 55.7\% & 9.0\% & 71 & \bb{381} \\
\hline
LP & 51.2\% & 72.5\% & \bb{21.6}\% & \bb{22.8}\% & 46 & 314 & LP & \bb{72.6}\% & 76.2\% & \bb{56.3}\% & 7.8\% & \bb{58} & 383 \\
DP1 & 51.4\% & 72.6\% & 19.2\% & 24.0\% & 34 & 280 & DP1 & 69.3\% & 75.6\% & 52.7\% & 7.8\% & 124 & 474 \\
DP2 & \bb{51.8}\% & 72.5\% & 20.4\% & 23.4\% & 38 & 295 & DP2 & 71.0\% & \bb{76.2}\% & 55.7\% & \bb{7.8}\% & 104 & 415 \\
\hline \hline
\multicolumn{7}{c ||}{Benchmark on Cyclist, DPM detections} & \multicolumn{7}{c |}{Benchmark on Cyclist, Regionlet detections} \\
\hline
Method & MOTA & MOTP & MT & ML & IDSW & FRAG & Method & MOTA & MOTP & MT & ML & IDSW & FRAG \\
\hline
SSP & 52.2\% & \bb{79.7}\% & 32.4\% & 29.7\% & 5 & \bb{11} & SSP & \bb{84.9}\% & \bb{82.3}\% & 73.0\% & 2.7\% & \bb{7} & \bb{18} \\
\hline
LP & \bb{57.2}\% & 79.5\% & \bb{43.2}\% & \bb{27.0}\% & 9 & 18 & LP & 83.2\% & 82.2\% & \bb{78.4}\% & 2.7\% & 10 & 22 \\
DP1 & 56.5\% & 79.4\% & 29.7\% & 32.4\% & 6 & 13 & DP1 & 80.1\% & 81.8\% & 70.3\% & 2.7\% & 12 & 29 \\
DP2 & 56.8\% & 79.6\% & 29.7\% & 32.4\% & \bb{5} & 12 & DP2 & 82.0\% & 82.2\% & 73.0\% & \bb{2.7}\% & 13 & 24 \\
\hline
\end{tabular}}
\end{center}
\vspace{0.05in}
\caption{Benchmark and cross-validation results on KITTI data set. Gray background
indicates the method uses target-specific appearance model. We evaluate two different detectors, Deformable Part
Models (DPM) and Regionlet, and different inference models with linear (SSP) and quadratic (LP,DP1,DP2)
cost models, each trained using SSVM.
}
\label{tab:tab2}
\end{table*}

\begin{table*}[tbp]
\begin{center}
\renewcommand{\tabcolsep}{3.0pt}
{\small
\begin{tabular}{ r | c | c | c |}
\multicolumn{4}{c}{MOTA on individual MOT test sequences} \\ \hline
& \cite{PirsiavashRF_CVPR_2011} & \cite{Milan:2016:PAMI} & Ours(LP) \\ \hline
ETH-Jelmoli &29.1 & 30.2 & 39.5 \\
ETH-Crossing & 20.0 & 16.5 & 24.9 \\
ETH-Linthescher & 15.9 & 17.0 & 15.6 \\
KITTI-16 & 23.2 & 34.0 & 39.2 \\
KITTI-19 & 8.3 & 17.4 & 28.2 \\
TUD-Crossing & 48.6 & 57.3 & 60.0 \\
ADL-Rundle-1 & -3.7 & 10.0 & 14.0\\
Venice-1 & 10.9 & 13.1 & 17.8 \\
PETS09-S2L2 & 33.8 & 37.5 & 41.5 \\
AVG-TownCentre & 6.6 & 8.2 & 14.7 \\
ADL-Rundle-3 & 12.8 & 16.9 & 28.0 \\ \hline
\end{tabular}
}
\end{center}
\vspace{0.05in}
\caption{Per-sequence accuracy comparison against purely motion-based algorithms
on MOT test-set.}
\label{tab:tab3}
\end{table*}

\begin{table*}[tbp]
\begin{center}
\renewcommand{\tabcolsep}{3.0pt}
{\small
\begin{tabular}{ r | c | c | c | c | c | c | c |}
\multicolumn{6}{c}{MOTA on other sequences} \\ \hline
& \cite{BrauICCV2013}* & \cite{Milan:2013:DTE} & \cite{ChariCVPR15} &
Ours(SSP) & Ours(LP) & Ours(SSP+Overfit) & Ours(LP+Overfit) \\ \hline
TUD-Stadtmitte & 70.0 & 56.2 & 51.6 & 48.1 & 48.6 & 46.6 & 49.0 \\
PETS09-S2.L1 & 83.0 & 90.3 & 85.5 & 83.3 & 83.5 & 85.8 & 86.2 \\
PETS09-S2.L2 & NA & 58.1 & 50.4 & 46.0 & 50.7 & 47.9 & 52.8 \\
PETS09-S2.L3 & NA & 39.8 & 40.3 & 40.7 & 41.3 & 40.3 & 41.3 \\
PETS09-S1.L1-2  & NA & 60.0 & 62.0& 57.2 & 59.9 & 57.2 & 59.4 \\
PETS09-S1.L2-1 & NA & 29.6 & 32.2 & 26.9 & 27.5 & 25.6 & 26.6 \\
\hline
Accumulated & NA & 55.4 & 53.9 & 49.8 & 51.6 & 50.1 & 52.6 \\
\hline
\end{tabular}
}
\end{center}
\vspace{0.05in}
\caption{Results on PETS and TUD-Stadtmitte. An asterisk(*)
indicates that the method uses different detector, training data and ground-truth than
other methods. SSP and LP corresponding to cross-validation results of linear/quadratic
models. +Overfit indicates the models are trained using all six sequences,
thus overfiting to the training data.
}
\label{tab:tab4}
\end{table*}

\subsection{Tracking Benchmark Results}
We start by comparing our model with various state-of-the-art results on the
three datasets.  Most competing methods on these datasets model high order
dynamics of either motion or appearance, or both, while our model uses very
simple motion model to build transition links, and do not explicitly employ any
appearance affinity model.

\paragraph{The MOT Benchmark}
For the MOT Benchmark, we only use a subset of contextual features that
includes the overlap and near relationships due to the varying view angle of
benchmark videos. Surprisingly, on MOTA score alone, we outperform many
state-of-the-art works without employing any explicit appearance/motion
model. We expect this is not because appearance/motion features are useless but
rather that the parameters of these features have not been optimally
learned/integrated into competing tracking methods.

\paragraph{The KITTI Tracking Benchmark}\footnote{In a recent update of the benchmark
server, the organizers changed their evaluation script to count detections in ``don't care''
regions as false positives, which we believe is not consistent with general consensus
of what ``don't care'' regions mean. Thus we report the results up to 05/24/2016 which
were evaluated using old evaluation script.}
Due to the high-speed motion of vehicle platforms, for the KITTI dataset we use
pre-computed frame-wise optical flow~(\cite{Liu2009}) to predict candidate detection
locations in future frames in order to generate candidate transition links between frames.
We evaluated two different detectors, DPM and the regionlets detector
(\cite{Wang_2013_ICCV}) which produced the best result in terms of MOTA, IDs and
FRAG during cross-validation. Results on the benchmark test set are summarized in
the upper part of Table~\ref{tab:tab2}. Notice that for cars on regionlet detection
set, we achieve almost equivalent MOTA score to that of \cite{ChoiICCV15}
which employs a novel flow descriptor, explicit high order dynamics and even
inter-trajectory interactions. 

\paragraph{PETS and TUD-Stadtmitte}
Similar to the MOT benchmark we use only a subset of contextual features. Since
cameras are fixed for all sequences in this setup, we take advantage of a
Kalman filter to predict object's position in future frames and build
transition links accordingly. The results are reported in Table~\ref{tab:tab4}.
Despite the fact that our model is relatively simple, we still achieve
comparable or better accuracy than state-of-the-art methods on most sequences.
The work of~\cite{ChariCVPR15} performs well on these sequences but we note
that it utilizes both body and head detectors to aid detection of heavily
overlapping pedestrians.
Interestingly, the model of~\cite{Milan:2013:DTE} outperforms both our model
and ~\cite{ChariCVPR15}, while on MOT and KITTI our model achieves a much
better MOTA than their Discrete-Continuous optimization framework (+5.6\% on
MOT, +10.3\% on KITTI-car and +5.9\% on KITTI-pedestrian). 

\subsection{Diagnostic Analysis}
\label{sec:diagnostic}
We conduct cross-validation experiments on the training set for MOT and KITTI benchmarks to study
the effect of quadratic terms, loss function and inference algorithm. The results
are summarized in Table \ref{tab:tab1} and \ref{tab:tab2}.  As shown in right
side of Table \ref{tab:tab1}, our novel loss function is superior to
traditional Hamming loss in terms of maximizing MOTA. The 1-pass DP proposed
in section \ref{sec:inference} achieves up to 10x speedup with
negligible loss in most metrics; 2-pass DP performs better than 1-pass DP
in most metrics, while still being up to 3x faster than LP
inference (Figure \ref{fig:fig6}) on long video sequence with dense objects.

We found SSP (min-cost flow without quadratic terms) achieves slightly better
overall accuracy on the MOT dataset. MOT only contains a single object category
and includes videos from many different viewpoints (surveillance, vehicle,
street level) which limits the potential benefits of simple 2D context
features.  However, by properly learning the detector confidence and transition
smoothness in the SSP model, many false tracks can be pruned even without
contextual knowledge.

For traditional multi-category detector such as DPM, quadratic interactions
were very helpful to improve the tracking performance on KITTI; this is most
evident for tracking cyclist, as shown in Table \ref{tab:tab2}, where LP, DP1
and DP2 all achieve considerable improvement over the baseline linear
objective.

However, when we switch to the much more accurate regionlet detector on KITTI,
LP inference achieves only slightly better results than SSP on car and pedestrian,
while losing to SSP on cyclist category. This is very similar to the result
on MOT dataset, where the LP and SSP models achieve almost equivalent results.
We attribute this to the increasingly accurate regionlet detector squeezing out any
relative gains to be had from our simple quadratic interactions. Interestingly, the 
gap between LP and the 1-pass DP approximation is also larger. Since the regionlet 
detector can often find objects with extreme occlusion and truncation, the tracking graph 
can become quite complicated and using one-shot greedy decisions for tracks can lead 
to inferior tracking result.  Two-pass DP, with its ability to ``fix'' potential errors
from previously found tracks, outperforms 1-pass DP by a noticeable margin in this scenario.

\subsection{Integrality Gap for Greedy Dynamic Programming}
We compare the accumulated cost from DP1, DP2, LP and the relaxed optimum on 21
training sequences of KITTI dataset. The visual comparison, as well as the
exact numbers are reported in Fig~\ref{fig:fig6}. We note that the LP method
often produces integral results even before rounding. This may be in part
because the the structured SVM will tend penalize fractional solution resulting
in learned model parameters that favors integral results.  On the other hand,
DP1 and DP2 which are greedy algorithms, do not benefit from this specific
property of structured SVM but they still manage to find good approximation
within 1\% of the relaxed optimum.

We note that there are two ways in which the ``greedy'' dynamic programming
can be suboptimal.  First, without quadratic interactions, the dynamic
programming approach only approximately optimizes the min-cost flow objective.
This optimality gap only appears when tracks that would have been found
correctly by SSP cross in the video sequence.  When there are no crossings,
1-pass greedy DP will find globally optimal solution, and in most real-world
tracking scenarios, 2-pass greedy DP is sufficient to fix most of the “errors”
1-pass DP could have made.

Second, for models with quadratic interactions, the greedy DP approach may
instance a single track when it might have been better (lower cost) to instance
a pair of tracks which have negative or zero interactions with each other while
each having strong positive interactions with the greedily instanced track
(remember we are minimizing the objective). Greedy selection algorithms have
some optimality guarantees in the case of submodular set functions (\ie
strictly positive pairwise interactions for the minimization problem). However,
the theoretical bounds are quite loose.  Empirically, we observe that the
integrality gap between the LP relaxation and integral solutions produced by
greedy DP are often within 1\% of the relaxed optimum, as shown in
Fig~\ref{fig:fig6}.

\begin{figure}[t]
 \begin{center}
 \begin{tabular}{cc}
 \includegraphics[clip,trim=0cm 0.25cm 0cm 0cm,width=0.47\linewidth]{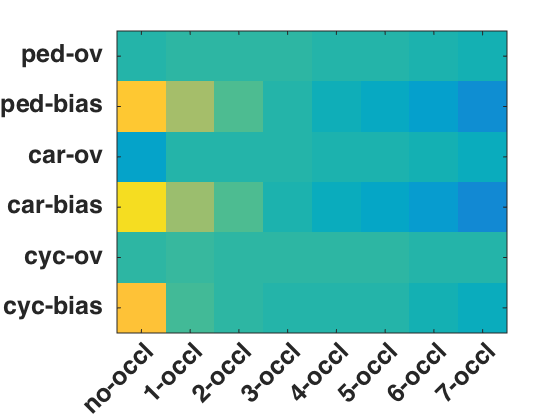}&
 \includegraphics[clip,trim=0cm 0.25cm 1cm 0cm,width=0.47\linewidth]{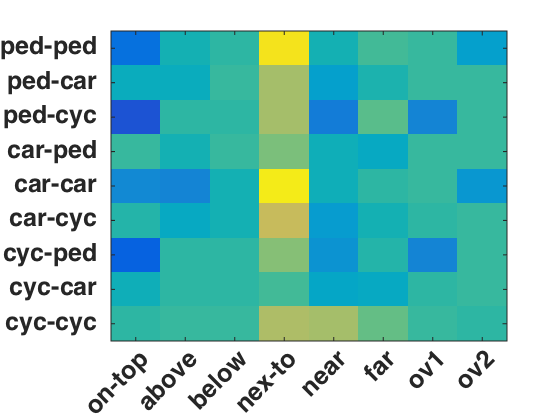}\\
 (a) inter-frame weights & (b) intra-frame weights\\
 \end{tabular}
\end{center}
   \caption{Visualization of the weight vector learned on KITTI dataset for 
   the DPM detector. Yellow has small cost, blue has large cost.
   (a) shows transition weights for different length of frame jumps. (b)
   shows learned pairwise contextual weights between objects.  The model
   encourages intra-class co-occurrence when objects are close but penalizes
   overlap and objects on top of others.  Note the strong negative interaction
   learned between cyclist and pedestrian (two classes which are easily
   confused by their respective detectors.). Figure is best
   viewed in color.}
   \label{fig:fig5}
\end{figure}
\begin{figure}[t]
 \begin{center}
 \begin{tabular}{cc}
 \includegraphics[clip,trim=0.1cm 0cm 1.8cm 0cm,width=0.21\textwidth]{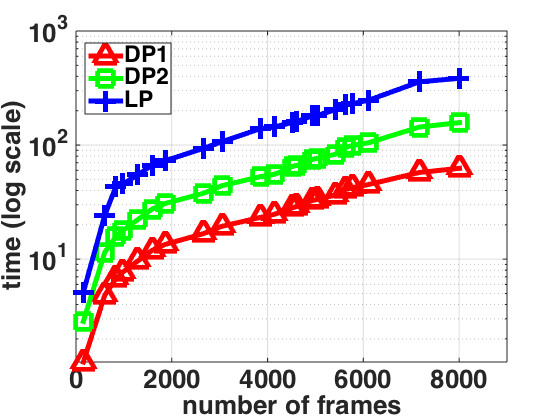}&
 \includegraphics[clip,trim=0.1cm 0cm 1.8cm 0cm,width=0.21\textwidth]{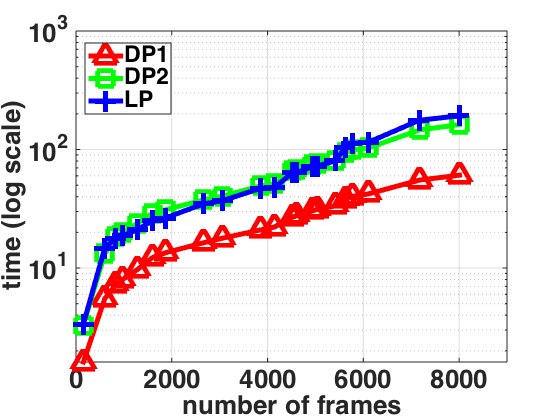}\\
 (a) KITTI-DPM Speed & (b) KITTI-Regionlet Speed \\
 \includegraphics[clip,trim=0.1cm 0cm 1.8cm 0cm,width=0.21\textwidth]{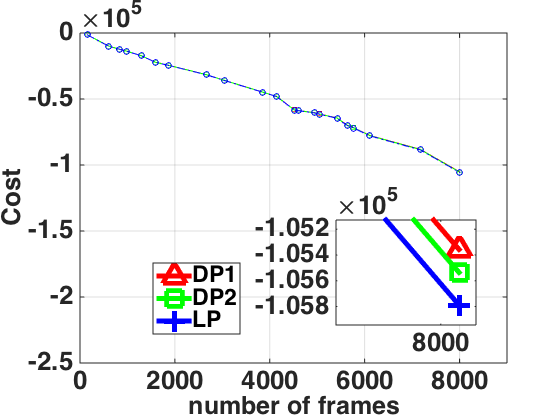}&
 \includegraphics[clip,trim=0.1cm 0cm 1.8cm 0cm,width=0.21\textwidth]{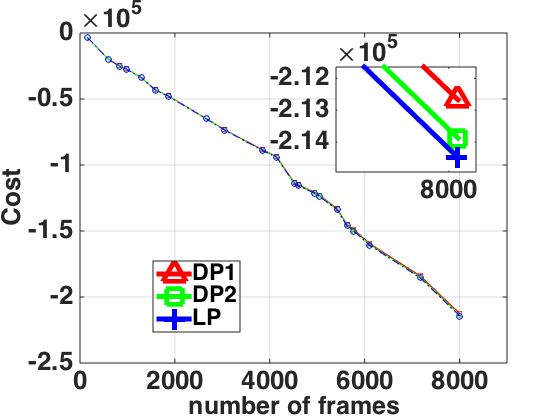}\\
 (c) KITTI-DPM Cost & (d) KITTI-Regionlet Cost \\
 &
 \end{tabular}
 \begin{tabular}{| c | c | c | c |}
 \hline
 \multicolumn{4}{| c |}{DPM detections} \\
 \hline
 Method & Time(seconds) & Time(FPS) & Cost \\
 \hline
 DP1 & 62.5 & 128.1 & -105356.636294 \\
 DP2 & 157.4 & 50.9 & -105535.299785 \\
 LP  & 384.9 & 20.8 & -105788.159640 \\
 Relax & NA & NA & -105788.871516 \\
 \hline \hline
 \multicolumn{4}{| c |}{Regionlet detections} \\
 \hline
 Method & Time(seconds) & Time(FPS) & Cost \\
 \hline
 DP1 & 60.7 & 131.9 & -212669.923615 \\
 DP2 & 162.4 & 49.3 & -213873.592597 \\
 LP  & 192.7 & 41.6 & -214483.825168 \\
 Relax & NA & NA & -214484.227818 \\
 \hline
 \multicolumn{4}{c}{(e) Detailed running time and cost}
 \end{tabular}
\end{center}
   \caption{Speed and quality comparison of proposed DP and traditional LP
   approximation over 21 sequences of KITTI training set. (a) and (b) are
   running time comparisons. Both DP1 and DP2 run much faster than LP inference,
	with DP1 being up to 10x faster on specific videos. (c) and (d)
   are cost comparisons. Notice that DP2 always finds lower total cost than
   DP1, which often results in better tracking performance (see
   Section~\ref{sec:diagnostic} for details). Figures are best viewed in color.
   Running times are averaged over three separate runs of each instance.}
   \label{fig:fig6}
\end{figure}

\begin{figure*}[t]
\begin{center}
\begin{tabular}{cc}
\includegraphics[clip,trim=0cm 0cm 0cm 0cm,width=0.47\textwidth,height=0.27\textwidth]{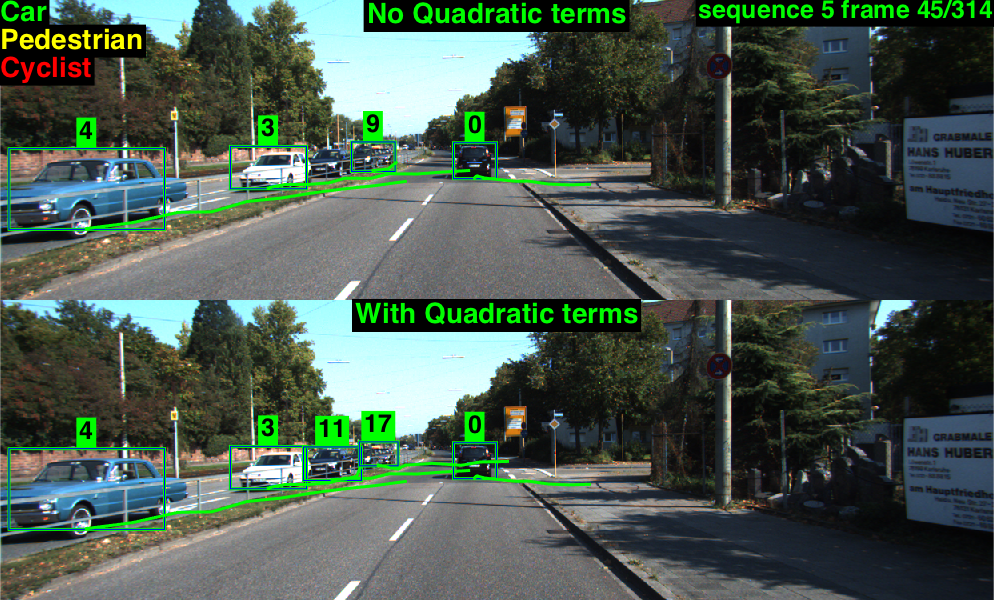}&
\includegraphics[clip,trim=0cm 0cm 0cm 0cm,width=0.47\textwidth,height=0.27\textwidth]{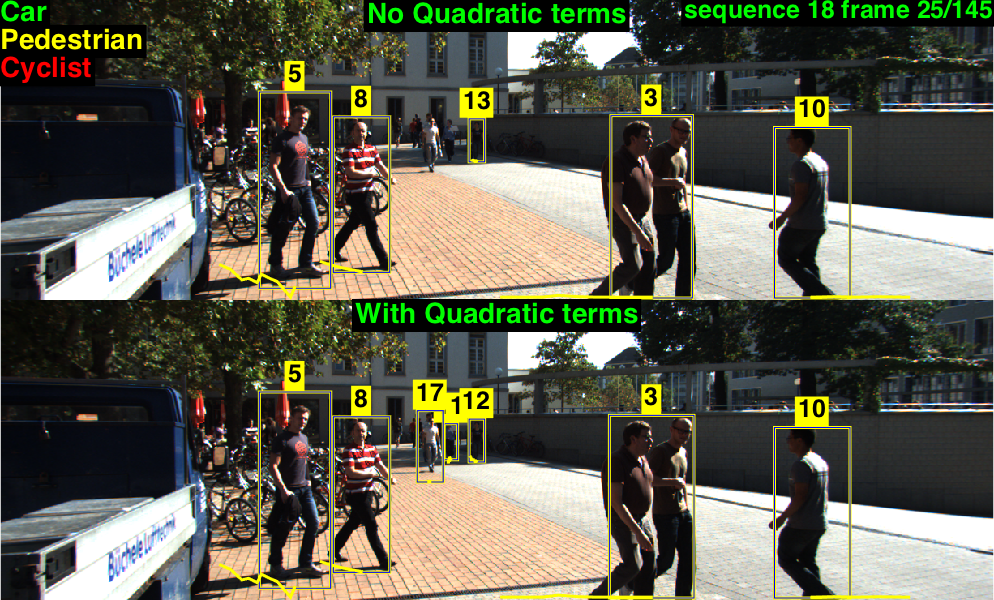}
\end{tabular}
\caption{Additional Qualitative Results on KITTI DPM detection set. With help of intra-class
quadratic interactions, the tracker is able to instantiate more correct trajectories.}
\end{center}
\end{figure*}

\begin{figure*}[t]
\begin{center}
\begin{tabular}{cc}
\includegraphics[clip,trim=0cm 0cm 0cm 0cm,width=0.47\textwidth,height=0.27\textwidth]{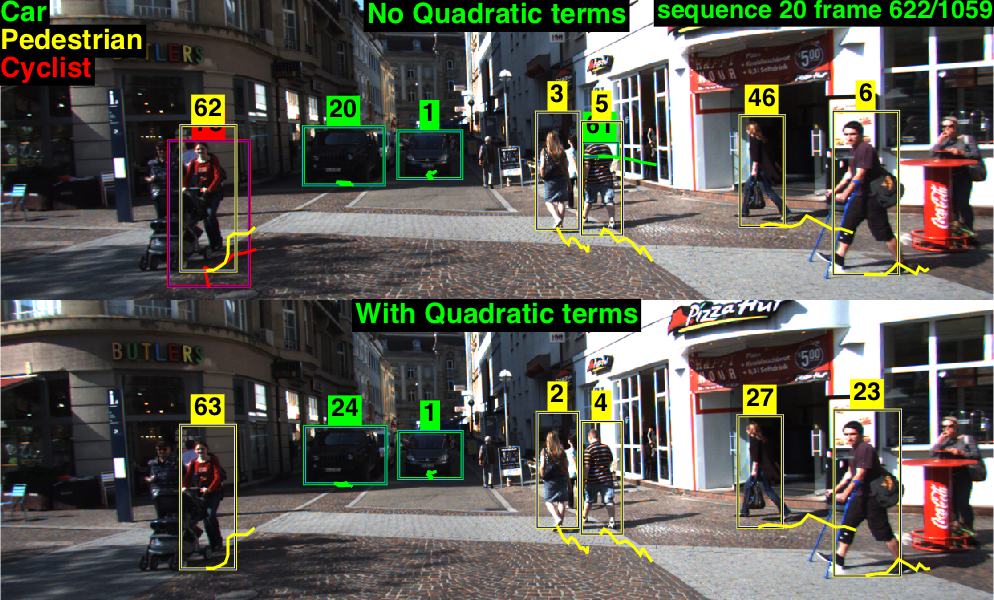}&
\includegraphics[clip,trim=0cm 0cm 0cm 0cm,width=0.47\textwidth,height=0.27\textwidth]{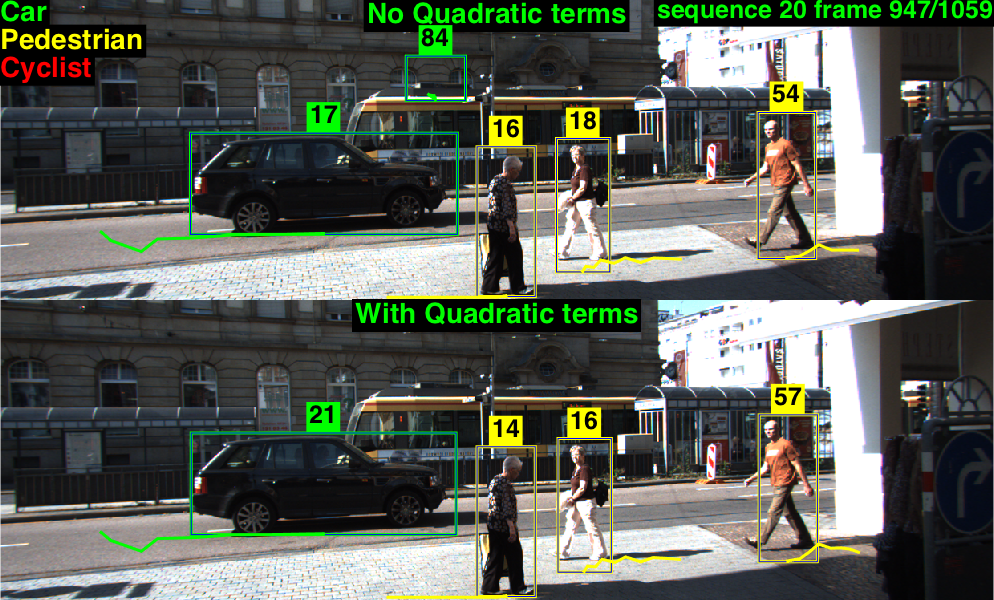}
\end{tabular}
\caption{Additional Qualitative Results on KITTI DPM detection set. The linear cost tracking model 
has wrongly labeled ``a person pushing a cart'' as ``cyclist'' (top-left), while our quadratic model is 
able to suppress cyclist trajectory with a high penalty between co-occurrence of pedestrian and cyclist
(bottom-left). The quadratic interaction is also helpful in that it could help to suppress spatially
infeasible co-occurrence from imperfect detectors, such as a car appearing on the back of a pedestrian
 (left), or a car ``flying'' above a pedestrian (right).
}
\end{center}
\end{figure*}

\begin{figure*}[t]
\begin{center}
\begin{tabular}{cc}
\includegraphics[clip,trim=0cm 0cm 0cm 0cm,width=0.35\textwidth,height=0.5\textwidth]{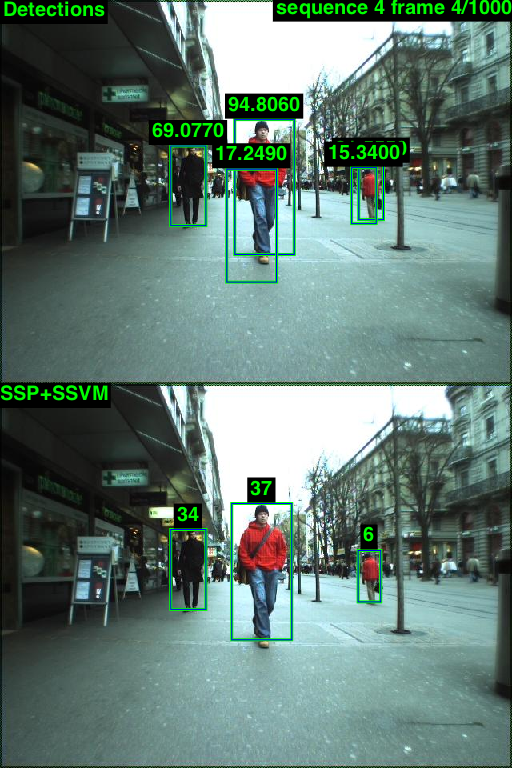}&
\includegraphics[clip,trim=0cm 0cm 0cm 0cm,width=0.35\textwidth,height=0.5\textwidth]{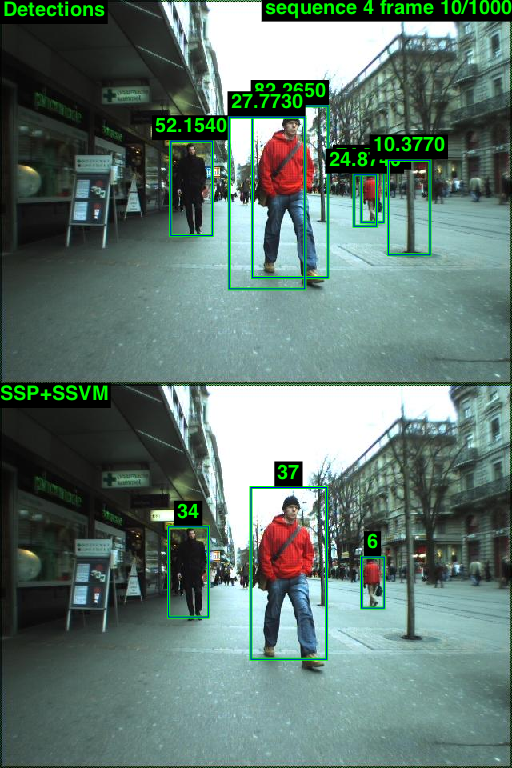}
\end{tabular}
\caption{By learning a proper parameter set, even a network-flow model without
pairwise potentials can prune away many false tracks by reasoning about detection confidence
and transition smoothness.
}
\end{center}
\end{figure*}

\subsection{Running Time}
We show running time comparison among DP1, DP2 and LP on KITTI training set
in Figure~\ref{fig:fig6}. For LP we only count the time spent inside
MOSEK library for relaxed inference, Frank-Wolfe rounding, plus the time for
backtracking final tracks. For DP1 and DP2 we count the entire
period spent inside our MATLAB function. On the DPM detection set, both
DP1 and DP2 run much faster than LP, with DP1 being up to 10x faster and
DP2 up to 3x faster. On regionlet detection set, the gap between DP2 and
LP becomes smaller, this is most likely due to the fact that regionlet
detection set has fewer candidate detections per sequence than DPM set,
and the overhead of MATLAB implementation comes to dominate. In fact, for
sequence 17 and sequence 20, on which both DP2 and LP take more than 10
seconds to finish using regionlet detections, DP2 runs 2x and 1.5x
faster than LP, respectively.

We note that our current implementation is not heavily optimized.  For example,
the dynamic programming step in our DP2 implementation only takes up to 
50\% of the total running time while contextual updates take up the remainder.
This suggests there is plenty of room to further accelerate and optimize the
DP algorithms.

\section{Conclusion and Future Work}
We have described algorithms for multi-target track association with quadratic 
interactions that are a natural extension of previously published approaches~(\cite{NetworkFlow,PirsiavashRF_CVPR_2011}).  Surprisingly,
the resulting system is able to outperform many far more complex state-of-the-art 
methods on both MOT
and KITTI benchmarks.  In contrast, simple application of the DP-based
tracker described by \cite{PirsiavashRF_CVPR_2011} does quite poorly on these
datasets (\eg, MOTA=14.9 on the MOT benchmark).  We attribute the
performance boost to our learning framework which produces much better
parameters than those estimated by hand-tuning or piece-wise model training.

A basic assumption of the network-flow models is that the entire video is available.
However it can become online by simply solving a new network-flow problem every time
we get a new frame; this might sound expensive at first, but remember that we can use the
caching strategy described in Section~\ref{sec:caching} to save the effort of computing
the first pass of dynamic programming, which is often the most time-consuming one.
\cite{LenzICCV2015} describes an online successive-shortest-path algorithm with fixed
size of memory, in which they fix the solution for nodes beyond a certain number of
frames. Obviously, our DP algorithms can do the same to achieve online inference with
bounded memory.

We stress that the ideas described here is also complimentary to other existing
methods.  While we did not see significant benefits to adding
simple appearance-based affinity features (e.g., RGB histogram or HOG) to our
model, many state-of-the-art systems perform hierarchical or streaming
data association which involves collecting examples from extended period of
trajectory to train target specific appearance models in an online fashion.
Such improved appearance models can be adapted to our framework, providing 
a way to explore more complicated affinity features while estimating hyper-parameters
automatically from data.  One could also introduce richer, trajectory level
contextual features under such a hierarchical learning framework.

\bibliographystyle{spbasic}      
\bibliography{egbib}   

\begin{thebibliography}{49}
\providecommand{\natexlab}[1]{#1}
\providecommand{\url}[1]{{#1}}
\providecommand{\urlprefix}{URL }
\expandafter\ifx\csname urlstyle\endcsname\relax
  \providecommand{\doi}[1]{DOI~\discretionary{}{}{}#1}\else
  \providecommand{\doi}{DOI~\discretionary{}{}{}\begingroup
  \urlstyle{rm}\Url}\fi
\providecommand{\eprint}[2][]{\url{#2}}

\bibitem[{Ahuja et~al(1993)Ahuja, Magnanti, and Orlin}]{SSP}
Ahuja RK, Magnanti TL, Orlin JB (1993) Network Flows: Theory, Algorithms, and
  Applications. Prentice-Hall, Inc., Upper Saddle River, NJ, USA

\bibitem[{Bae and Yoon(2014)}]{Bae_2014_CVPR}
Bae SH, Yoon KJ (2014) Robust online multi-object tracking based on tracklet
  confidence and online discriminative appearance learning. In: Proc. of CVPR

\bibitem[{Bernardin and
  Stiefelhagen(2008)}]{Bernardin:2008:EMO:1384968.1453688}
Bernardin K, Stiefelhagen R (2008) Evaluating multiple object tracking
  performance: The clear mot metrics. Journal on Image Video Processing
  \doi{10.1155/2008/246309}

\bibitem[{Brau et~al(2013)Brau, Guan, Simek, Del~Pero, Reimer~Dawson, and
  Barnard}]{BrauICCV2013}
Brau E, Guan J, Simek K, Del~Pero L, Reimer~Dawson C, Barnard K (2013) Bayesian
  3d tracking from monocular video. In: Proc. of ICCV

\bibitem[{Brendel et~al(2011)Brendel, Amer, and
  Todorovic}]{Brendel11multiobjecttracking}
Brendel W, Amer M, Todorovic S (2011) Multiobject tracking as maximum weight
  independent set. In: Proc. of CVPR

\bibitem[{Butt and Collins(2013)}]{Butt_2013_ICCV_Workshops}
Butt AA, Collins RT (2013) Multi-target tracking by lagrangian relaxation to
  min-cost network flow. In: Proc. of CVPR

\bibitem[{Chari et~al(2015)Chari, Lacoste-Julien, Laptev, and
  Sivic}]{ChariCVPR15}
Chari V, Lacoste-Julien S, Laptev I, Sivic J (2015) On pairwise costs for
  network flow multi-object tracking. In: Proc. of CVPR

\bibitem[{Choi(2015)}]{ChoiICCV15}
Choi W (2015) Near-online multi-target tracking with aggregated local flow
  descriptor. In: Proc. of ICCV

\bibitem[{Choi and Savarese(2012)}]{choi_eccv12}
Choi W, Savarese S (2012) A unified framework for multi-target tracking and
  collective activity recognition. In: Proc. of ECCV

\bibitem[{Dehghan et~al(2015)Dehghan, Tian, Torr, and Shah}]{DehghanCVPR2015}
Dehghan A, Tian Y, Torr PH, Shah M (2015) Target identity-aware network flow
  for online multiple target tracking. In: Proc. of CVPR

\bibitem[{Desai et~al(2009)Desai, Ramanan, and Fowlkes}]{DesaiRF_ICCV_2009}
Desai C, Ramanan D, Fowlkes C (2009) Discriminative models for multi-class
  object layout. In: Proc. of ICCV

\bibitem[{Doll\'ar et~al(2014)Doll\'ar, Appel, Belongie, and
  Perona}]{DollarPAMI14pyramids}
Doll\'ar P, Appel R, Belongie S, Perona P (2014) Fast feature pyramids for
  object detection. IEEE Transactions on Pattern Analysis and Machine
  Intelligence 36(8):1532--1545

\bibitem[{Felzenszwalb et~al(2010)Felzenszwalb, Girshick, McAllester, and
  Ramanan}]{voc-release4}
Felzenszwalb PF, Girshick RB, McAllester D, Ramanan D (2010) Object detection
  with discriminatively trained part based models. IEEE Transactions on Pattern
  Analysis and Machine Intelligence 32(9):1627 -- 1645

\bibitem[{Finley and Joachims(2008)}]{Finley/Joachims/08a}
Finley T, Joachims T (2008) Training structural {SVMs} when exact inference is
  intractable. In: Proc. of ICML

\bibitem[{Geiger et~al(2012)Geiger, Lenz, and Urtasun}]{Geiger2012CVPR}
Geiger A, Lenz P, Urtasun R (2012) Are we ready for autonomous driving? the
  kitti vision benchmark suite. In: Proc. of CVPR

\bibitem[{Geiger et~al(2013)Geiger, Lenz, Stiller, and
  Urtasun}]{Geiger2013IJRR}
Geiger A, Lenz P, Stiller C, Urtasun R (2013) Vision meets robotics: The kitti
  dataset. International Journal of Robotics Research 32(11):1231--1237

\bibitem[{Geiger et~al(2014)Geiger, Lauer, Wojek, Stiller, and
  Urtasun}]{Geiger2014PAMI}
Geiger A, Lauer M, Wojek C, Stiller C, Urtasun R (2014) 3d traffic scene
  understanding from movable platforms. IEEE Transactions on Pattern Analysis
  and Machine Intelligence 36(5):1012 -- 1025

\bibitem[{Joachims et~al(2009)Joachims, Finley, and Yu}]{Joachims/etal/09a}
Joachims T, Finley T, Yu CN (2009) Cutting-plane training of structural svms.
  Machine Learning 77(1):27--59

\bibitem[{Joulin et~al(2014)Joulin, Tang, and Fei-Fei}]{TangECCV14}
Joulin A, Tang K, Fei-Fei L (2014) Efficient image and video co-localization
  with frank-wolfe algorithm. In: Proc. of ECCV

\bibitem[{Kim et~al(2015)Kim, Li, Ciptadi, and Rehg}]{KimICCV15}
Kim C, Li F, Ciptadi A, Rehg JM (2015) Multiple hypothesis tracking revisited.
  In: Proc. of ICCV

\bibitem[{Kim et~al(2013)Kim, Kwak, Feyereisl, and
  Han}]{Kim:2012:OMT:2482048.2482058}
Kim S, Kwak S, Feyereisl J, Han B (2013) Online multi-target tracking by large
  margin structured learning. In: Proc. of ACCV

\bibitem[{Lacoste-Julien et~al(2006)Lacoste-Julien, Taskar, Klein, and
  Jordan}]{Lacoste-Julien:2006:WAV:1220835.1220850}
Lacoste-Julien S, Taskar B, Klein D, Jordan MI (2006) Word alignment via
  quadratic assignment. In: Proc. of HLT-NAACL

\bibitem[{Leal-Taix{\'e} et~al(2014)Leal-Taix{\'e}, Fenzi, Kuznetsova,
  Rosenhahn, and Savarese}]{LeaFen2014}
Leal-Taix{\'e} L, Fenzi M, Kuznetsova A, Rosenhahn B, Savarese S (2014)
  Learning an image-based motion context for multiple people tracking. In:
  Proc. of CVPR

\bibitem[{Leal-Taix\'{e} et~al(2015)Leal-Taix\'{e}, Milan, Reid, Roth, and
  Schindler}]{MOTChallenge2015}
Leal-Taix\'{e} L, Milan A, Reid I, Roth S, Schindler K (2015) {MOTC}hallenge
  2015: {T}owards a benchmark for multi-target tracking. arXiv:150401942 [cs]

\bibitem[{Lenz et~al(2015)Lenz, Geiger, and Urtasun}]{LenzICCV2015}
Lenz P, Geiger A, Urtasun R (2015) Followme: Efficient online min-cost flow
  tracking with bounded memory and computation. In: Proc. of ICCV

\bibitem[{Li et~al(2009)Li, Huang, and Nevatia}]{Li09learningto}
Li Y, Huang C, Nevatia R (2009) Learning to associate: Hybridboosted
  multi-target tracker for crowded scene. In: Proc. of CVPR

\bibitem[{Liu(2009)}]{Liu2009}
Liu C (2009) Beyond pixels: Exploring new representations and applications for
  motion analysis. PhD thesis, Massachusetts Institute of Technology

\bibitem[{Lou and Hamprecht(2011)}]{lou_11_structured}
Lou X, Hamprecht FA (2011) Structured learning for cell tracking. In: Proc. of
  NIPS

\bibitem[{Milan et~al(2012)Milan, Schindler, and Roth}]{Andriyenko:2012:DCO}
Milan A, Schindler K, Roth S (2012) Discrete-continuous optimization for
  multi-target tracking. In: Proc. of CVPR

\bibitem[{Milan et~al(2013)Milan, Schindler, and Roth}]{Milan:2013:DTE}
Milan A, Schindler K, Roth S (2013) Detection- and trajectory-level exclusion
  in multiple object tracking. In: Proc. of CVPR

\bibitem[{Milan et~al(2014)Milan, Roth, and Schindler}]{Milan:2014:CEM}
Milan A, Roth S, Schindler K (2014) Continuous energy minimization for
  multitarget tracking. IEEE Transactions on Pattern Analysis and Machine
  Intelligence 36(1):58--72

\bibitem[{Milan et~al(2015)Milan, Leal-Taix{\'e}, Schindler, and
  Reid}]{Milan:2015:CVPR}
Milan A, Leal-Taix{\'e} L, Schindler K, Reid I (2015) Joint tracking and
  segmentation of multiple targets. In: Proc. of CVPR

\bibitem[{Milan et~al(2016)Milan, Schindler, and Roth}]{Milan:2016:PAMI}
Milan A, Schindler K, Roth S (2016) Multi-target tracking by
  discrete-continuous energy minimization. IEEE Transactions on Pattern
  Analysis and Machine Intelligence 38(10):2054 -- 2068

\bibitem[{Pirsiavash et~al(2011)Pirsiavash, Ramanan, and
  Fowlkes}]{PirsiavashRF_CVPR_2011}
Pirsiavash H, Ramanan D, Fowlkes CC (2011) Globally-optimal greedy algorithms
  for tracking a variable number of objects. In: Proc. of CVPR

\bibitem[{Segal and Reid(2013)}]{SegalICCV2013}
Segal AV, Reid I (2013) Latent data association: Bayesian model selection for
  multi-target tracking. In: Proc. of ICCV

\bibitem[{Solera et~al(2015)Solera, Calderara, and Cucchiara}]{SoleraICCV2015}
Solera F, Calderara S, Cucchiara R (2015) Learning to divide and conquer for
  online multi-target tracking. In: Proc. of ICCV

\bibitem[{Szummer et~al(2008)Szummer, Kohli, and Hoiem}]{Szummer_2008_ECCV}
Szummer M, Kohli P, Hoiem D (2008) Learning crfs using graph cuts. In: ECCV

\bibitem[{Tang et~al(2013)Tang, Andriluka, Milan, Schindler, Roth, and
  Schiele}]{TangICCV2013}
Tang S, Andriluka M, Milan A, Schindler K, Roth S, Schiele B (2013) Learning
  people detectors for tracking in crowded scenes. In: Proc. of ICCV

\bibitem[{Tang et~al(2015)Tang, Andres, Andriluka, and Schiele}]{TangCVPR2015}
Tang S, Andres B, Andriluka M, Schiele B (2015) Subgraph decomposition for
  multi-target tracking. In: Proc. of CVPR

\bibitem[{Taskar et~al(2003)Taskar, Guestrin, and Koller}]{Taskar03M3N}
Taskar B, Guestrin C, Koller D (2003) Max-margin markov networks. In: Proc. of
  NIPS

\bibitem[{Wang et~al(2014)Wang, Wang, Luk~Chan, and Wang}]{Wang_2014_CVPR}
Wang B, Wang G, Luk~Chan K, Wang L (2014) Tracklet association with online
  target-specific metric learning. In: Proc. of CVPR

\bibitem[{Wang and Fowlkes(2015)}]{WangF_BMVC_2015}
Wang S, Fowlkes CC (2015) Learning optimal parameters for multi-target
  tracking. In: Proc. of BMVC

\bibitem[{Wang et~al(2013)Wang, Yang, Zhu, and Lin}]{Wang_2013_ICCV}
Wang X, Yang M, Zhu S, Lin Y (2013) Regionlets for generic object detection.
  In: Proc. of ICCV

\bibitem[{Wu et~al(2012)Wu, Thangali, Sclaroff, , and Betke}]{WuThScBe2012}
Wu Z, Thangali A, Sclaroff S, , Betke M (2012) Coupling detection and data
  association for multiple object tracking. In: Proc. of CVPR

\bibitem[{Xiang et~al(2015)Xiang, Alahi, and Savarese}]{XiangICCV15}
Xiang Y, Alahi A, Savarese S (2015) Learning to track: Online multi-object
  tracking by decision making. In: Proc. of ICCV

\bibitem[{Yang and Nevatia(2012)}]{Yang12anonline}
Yang B, Nevatia R (2012) An online learned crf model for multi-target tracking.
  In: Proc. of CVPR

\bibitem[{Yoon et~al(2015)Yoon, Yang, Lim, and Yoon}]{DBLP:conf/wacv/YoonYLY15}
Yoon JH, Yang M, Lim J, Yoon K (2015) Bayesian multi-object tracking using
  motion context from multiple objects. In: Proc. of WACV

\bibitem[{Zaied and Shawky(2014)}]{ANasser14}
Zaied ANH, Shawky LAE (2014) A survey of quadratic assignment problems.
  International Journal of Computer Applications 101(6):28--36

\bibitem[{Zhang et~al(2008)Zhang, Li, and Nevatia}]{NetworkFlow}
Zhang L, Li Y, Nevatia R (2008) Global data association for multi-object
  tracking using network flows. In: Proc. of CVPR

\end{thebibliography}

\end{document}